\pdfoutput=1

\documentclass[11pt]{article}

\usepackage[preprint]{acl}

\usepackage{times}
\usepackage{latexsym}

\usepackage[T1]{fontenc}

\usepackage[utf8]{inputenc}

\usepackage{microtype}

\usepackage{inconsolata}

\usepackage{graphicx}
\usepackage{adjustbox}

\usepackage{minted}
\setminted{
  breaklines=true,
  rulecolor=\color{gray},
  frame=lines,
  framerule=0.5pt,
  framesep=3mm
}
\usepackage{etoolbox}
\BeforeBeginEnvironment{minted}{\vspace{10em}}
\usepackage{booktabs}
\usepackage{xspace}
\usepackage[dvipsnames]{xcolor}
\usepackage{pifont}
\usepackage{amsmath}
\usepackage{tabularx}
\usepackage{colortbl}
\definecolor{tableheader}{HTML}{EFEFEF}
\usepackage{multirow}
\usepackage{listings}
\usepackage{algorithm}
\usepackage{algpseudocode}
\usepackage{hyperref}
\usepackage{tikz}
\usetikzlibrary{shapes.geometric}
\usepackage[capitalize,nameinlink]{cleveref}
\usepackage{xcolor}
\usepackage{tabularx}
\usepackage{array}
\newcolumntype{Y}{>{\centering\arraybackslash}X}
\usepackage{xcolor}
\definecolor{tableheader}{RGB}{220,220,220}
\usepackage{float}
\usepackage{multirow}
\floatstyle{plain}     
\newfloat{prompt}{htbp}{lop}
\floatname{prompt}{Prompt}
\usepackage{listings}
\usepackage[most]{tcolorbox}
\usepackage{enumitem}
\usepackage{caption}
\usepackage{subcaption}
\usepackage{float} 

\tcbset{
  promptbox/.style={
    enhanced,
    breakable,
    colback=gray!5,
    colframe=black!60,
    boxrule=0.6pt,
    arc=2pt,
    left=6pt,
    right=6pt,
    top=6pt,
    bottom=6pt,
    fonttitle=\bfseries\footnotesize,
    coltitle=black,
    colbacktitle=gray!65,
    attach title to upper,
    before skip=1em,
    after skip=1em,

    overlay unbroken={},
    overlay first={},
    overlay middle={
      \node[
        anchor=south east,
        font=\scriptsize\itshape
        text=black!50

      ] at ([xshift=-6pt,yshift=4pt]frame.south east)
      {(continued on next page)};
    },
    overlay last={},
  }
}

\newcommand{\Teacher}{\mathcal{T}}
\newcommand{\Pruner}{\mathcal{P}}
\newcommand{\Student}{\mathcal{S}}

\newcommand{\thetaP}{\theta_{\Pruner}}

\newcommand{\legendmark}[4][solid]{%
  \tikz[baseline=-0.6ex]{
    \draw[#2, #1, line width=1.0pt] (-0.35,0) -- (0.35,0);

    \ifx#3\empty\else
      \node[
        draw=#2,
        fill=#2,
        shape=#3,
        minimum size=#4,
        inner sep=0pt
      ] at (0,0) {};
    \fi
  }%
}

\definecolor{symbolic}{RGB}{98, 160, 202}
\definecolor{grammar}{RGB}{255, 165, 85}
\definecolor{entity}{RGB}{107, 188, 107}
\definecolor{meta}{RGB}{226, 104, 104}
\definecolor{verbal}{RGB}{180, 148, 209}
\definecolor{coref}{RGB}{174, 136, 129}
\definecolor{ub}{RGB}{227, 119, 195}

\makeatletter
\renewcommand{\@fnsymbol}[1]{}
\makeatother

\title{
\textcolor{black!40}{Do}
\textcolor{black!100}{LLMs}
\textcolor{black!65}{Encode}
\textcolor{black!100}{Functional}
\textcolor{black!100}{Importance}
\textcolor{black!40}{of}
\textcolor{black!100}{Reasoning}
\textcolor{black!65}{Tokens}
\textcolor{black!40}{?}
}

\author{Janvijay Singh \qquad Dilek Hakkani-Tür\\
        The Grainger College of Engineering\\University of Illinois Urbana Champaign\\
        \texttt{\{jvsingh2, dilek\}@illinois.edu}}

\begin{document}
\maketitle
\begin{abstract}
Large language models solve complex tasks by generating long reasoning chains, achieving higher accuracy at the cost of increased computational cost and reduced ability to isolate functionally relevant reasoning.
Prior work on compact reasoning shortens such chains through probabilistic sampling, heuristics, or supervision from frontier models, but offers limited insight into whether models internally encode token-level functional importance for answer generation.
We address this gap diagnostically and propose \textit{greedy pruning}, a likelihood-preserving deletion procedure that iteratively removes reasoning tokens whose removal minimally degrades model likelihood under a specified objective, yielding length-controlled reasoning chains.
We evaluate pruned reasoning in a distillation framework and show that students trained on pruned chains outperform a frontier-model–supervised compression baseline at matched reasoning lengths.
Finally, our analysis reveals systematic pruning patterns and shows that attention scores can predict greedy pruning ranks, further suggesting that models encode a nontrivial functional importance structure over reasoning tokens.\footnote{
\fontsize{9.3pt}{11pt}\selectfont
\textbf{Code:} \href{https://github.com/iamjanvijay/greedy-token-pruner}{github.com/iamjanvijay/greedy-token-pruner};
\textbf{Demo:} \href{https://iamjanvijay.github.io/greedy-token-pruner/}{iamjanvijay.github.io/greedy-token-pruner}.
}
\end{abstract}
\section{Introduction}

Large Language Models (LLMs) rely on long, explicit reasoning chains to solve complex tasks in mathematics, science, and multi-step decision making~\cite{10.5555/3600270.3602070,ElKishky2024OpenAIOS,Guo2025DeepSeekR1IR}. 
While long reasoning chains improve performance, their length incurs substantial costs, including higher inference latency, increased training and memory requirements,
and greater difficulty in isolating parts of the reasoning that are functionally responsible for the final answer.
This trade-off has motivated a growing line of work on producing more efficient and compact reasoning chains that preserve task performance~\cite{10.5555/3737916.3739357,10.1609/aaai.v39i23.34608,aggarwal2025l,feng2025efficient}.
Most existing approaches to compact reasoning achieve compression by explicitly exploring the space of reasoning chains and selecting shorter alternatives.
One prominent mechanism is temperature sampling, in which methods generate multiple candidate chains and train models to produce shorter chains among them~\cite{Hassid2025DontOI,aggarwal2025l,Luo2025O1PrunerLF,anonymous2025thinkprune}.
Other mechanisms include LLM probability-based heuristics or post-processing rules for removing reasoning segments, as well as the use of frontier LLMs to generate candidate compact reasoning chains~\cite{10.1609/aaai.v39i23.34608,xia-etal-2025-tokenskip,qiao-etal-2025-concise}.
Although effective, these approaches provide limited insight into whether the reasoning LLM internally encodes
\textit{token-level functional importance} in its reasoning for answer generation.

Compared to prior work, we take a different perspective and ask a more fundamental question:
\textit{Do LLMs internally encode the functional importance of reasoning tokens for answer generation?}
Rather than proposing another method for generating compact reasoning, we study whether a model’s own likelihood and attention patterns can serve as signals for ranking reasoning tokens by their functional importance.
This framing treats reasoning compression as a diagnostic problem aimed at revealing internally encoded token-level functional structure in the model’s reasoning.

To this end, we introduce \textit{greedy pruning}, a likelihood-preserving deletion procedure inspired by perturbation-based attribution methods~\cite{10.1007/978-3-319-10590-1_53,zintgraf2017visualizing} and greedy decoding.
Perturbation-based attribution methods assess the importance of input parts by perturbing or removing them and observing the resulting changes in the model’s output, 
while greedy decoding in LLMs incrementally adds tokens to maximize likelihood.
Building on these ideas, greedy pruning starts from a complete reasoning chain and incrementally removes tokens whose deletion minimally degrades the model’s likelihood under a specific objective (described in Section~\ref{sec:greedy}).
This process produces a ranking over reasoning tokens and a set of length-controlled reasoning chains, where pruning ranks reflect functional importance under the model’s own distribution.
Under this interpretation, greedy pruning serves as a probe for identifying reasoning-token subsequences important for maintaining predictive behavior.

To test whether greedily pruned reasoning chains indeed preserve functionally important tokens, we evaluate them using a teacher–pruner–student distillation framework across multiple reasoning benchmarks, including GSM8K~\cite{Cobbe2021TrainingVT}, MATH~\cite{hendrycksmath2021}, and the multi-domain benchmark MMLU-Pro~\cite{wang2024mmlupro}.
Quantitatively, we show that students distilled on greedily pruned reasoning chains outperform multiple pruning baselines, including \texttt{TokenSkip}~\cite{xia-etal-2025-tokenskip}, which relies on token importance labels from frontier models.
Qualitatively, we analyze the functional structure of pruned reasoning chains, revealing consistent patterns in which types of tokens are preserved or removed at different pruning stages.
Finally, we examine whether pruning ranks can be predicted from attention scores, providing evidence that importance signals revealed by greedy pruning are accessible from the model’s own internals.

Overall, our results suggest that LLMs encode a nontrivial token-level functional structure within reasoning, which can be revealed through likelihood-preserving pruning and corroborated by attention-based signals.
Beyond reasoning compression, greedy pruning offers a principled tool for probing the internal organization of LLM-generated reasoning.

\section{Related Work}

\paragraph{Compact Reasoning in LLMs.} 
Recent work has explored reducing the costs of long reasoning chains by compressing model-generated reasoning sequences.
At a high level, most approaches discover compact chains through external exploration mechanisms and then train models to prefer shorter candidates.
A common strategy relies on stochastic sampling, where LLMs generate multiple reasoning chains using temperature-controlled decoding and are trained via supervised fine-tuning or reinforcement learning to favor shorter yet correct outputs~\cite{10.5555/3737916.3739357,Hassid2025DontOI,aggarwal2025l,Luo2025O1PrunerLF,anonymous2025thinkprune,yi2025shorterbetter}. 
Other methods apply LLM probability-based heuristics or post-processing rules to prune reasoning steps
~\cite{li-etal-2023-compressing,qiao-etal-2025-concise,Zeng2025PruningTU}.
A third line of work employs frontier LLMs to compress reasoning chains and distills smaller models on the resulting outputs~\cite{10.1609/aaai.v39i23.34608,Cui2025StepwisePR,xia-etal-2025-tokenskip}. 
Despite their differences, prior approaches share a common design choice: compact reasoning chains are discovered through external mechanisms such as stochastic exploration, heuristic rules, or additional supervision. 
Consequently, they offer limited controllability, provide little characterization of removed reasoning tokens, and shed limited light on whether token-level importance is encoded by the model. 
In contrast, our approach enables explicit and controllable pruning, directly characterizes removed reasoning tokens, and provides a principled diagnostic tool for probing internal token-level importance.

\begin{figure*}[t]
\centering
\includegraphics[width=1.0\linewidth]{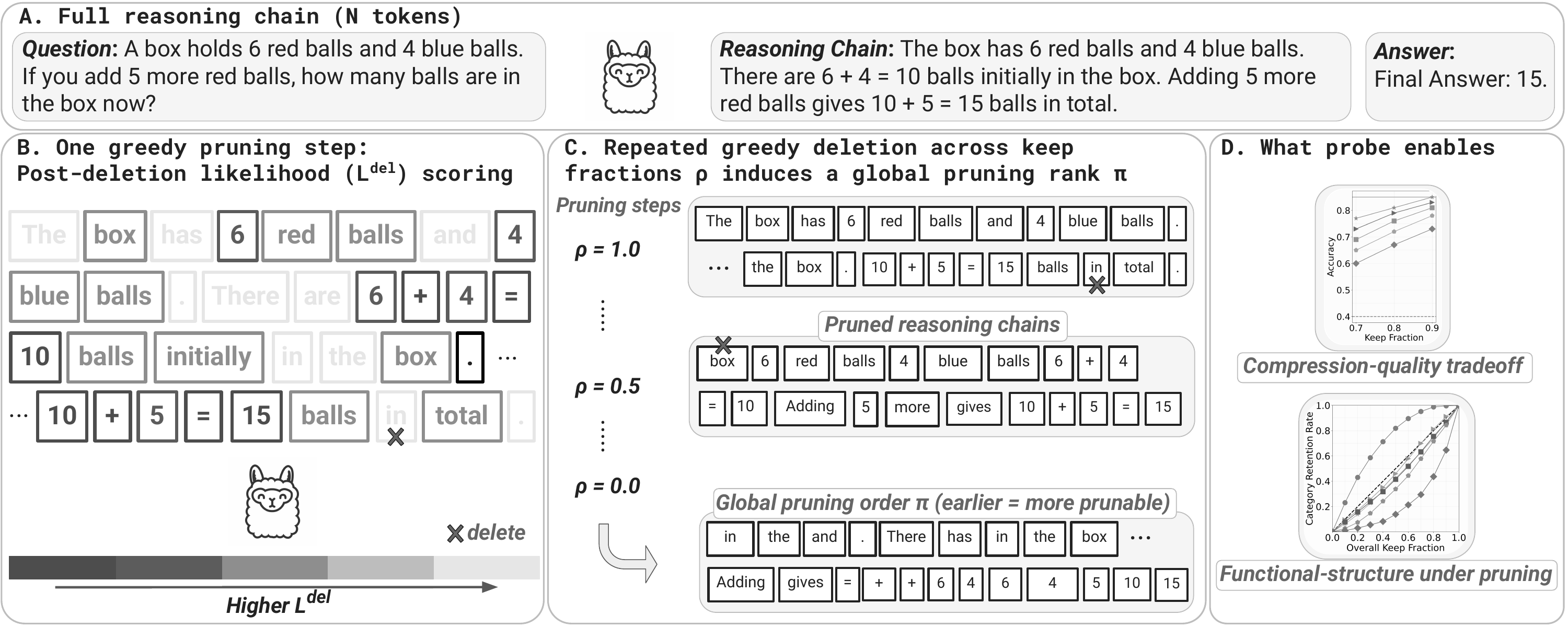}

\caption{
\textbf{Greedy pruning as a diagnostic probe.}
\textbf{A.} A teacher model generates a full reasoning chain for a given question.
\textbf{B.} A greedy pruning step scores candidate token deletions by post-deletion likelihood $L^{del}$ and removes the token whose deletion best preserves likelihood.
\textbf{C.} Iterating this procedure over decreasing keep fractions $\rho$ yields length-controlled chains and induces a pruning order $\pi$, where earlier-ranked tokens are safer to prune.
\textbf{D.} This order enables analysis of compression–quality tradeoffs and the structure of encoded functional importance.
}
\label{fig:main_figure}
\end{figure*}

\paragraph{Attribution Methods for LLM Reasoning.}
Attribution methods identify parts of the input or model responsible for a prediction by tracing importance signals through the network~\citep{10.1145/3639372}.
In neural language models, such signals have been derived from input gradients~\citep{10.5555/3305890.3306024,yin-neubig-2022-interpreting}, attention patterns~\citep{abnar-zuidema-2020-quantifying,Hou2023TowardsAM}, representation similarity~\citep{yin-neubig-2022-interpreting,enguehard-2023-sequential}, and perturbation-based deletion analyses~\citep{10.1007/978-3-319-10590-1_53,10.1145/2939672.2939778,zintgraf2017visualizing}. 
Recent work extends attribution to reasoning in LLMs, including Shapley-style analyses of reasoning tokens~\citep{gao2023shapleycot}, gradient-based importance scores~\citep{Wu2023AnalyzingCP}, intervention-based faithfulness tests~\citep{Lanham2023MeasuringFI}, attention-based attribution~\citep{CohenWang2025LearningTA}, and counterfactual resampling of reasoning steps~\citep{bigelow2025forking,anonymous2025thought}. 
These methods reveal that reasoning traces exhibit some internal importance structure, but are primarily restricted to analysis and understanding.
In contrast, our work adopts a more pragmatic perspective~\citep{nanda2025pragmaticinterp}, aiming to produce internal, token-level importance rankings for reasoning tokens that can also be used to train models for compact reasoning.

\section{Greedy Pruning: Likelihood-Preserving Deletion of Reasoning Tokens}
\label{sec:greedy_pruning}

\subsection{Problem Setup}
\label{sec:setup}

Given a question $Q$, we consider three model roles: (i) a teacher LLM $\Teacher$ that generates a reasoning chain $R=(r_1,\dots,r_n)$ of $n$ tokens and an answer $A$; (ii) a pruner LLM $\Pruner$ with parameters $\thetaP$ that evaluates likelihood-based objectives over subsequences of $R$; and (iii) a student model $\Student$ that can be trained on pruned reasoning chains.
We refer to this as a teacher–pruner–student distillation framework, which we instantiate in Section~\ref{sec:experiments}.
We study pruning in the \emph{post-generation} setting, where a complete reasoning chain is first produced and pruning is applied offline.
Given a target keep fraction $\rho\in(0,1]$, let $m=\lceil \rho n \rceil$ denote the retained length, and let $R_K$ denote the subsequence of $R$ indexed by the kept index set $K\subseteq\{1,\dots,n\}$.
Pruning is defined with respect to a likelihood-based objective $\mathcal{L}_{\thetaP}(Q,R_K,A)$ under the pruner model, with the goal of identifying a subset $K$ of size $m$ such that $\mathcal{L}$ is approximately preserved. 

\subsection{Greedy Pruning}
\label{sec:greedy}

Greedy pruning is a likelihood-preserving deletion procedure, analogous to greedy decoding. 
While greedy decoding incrementally \emph{adds} tokens to maximize likelihood, greedy pruning starts from a complete reasoning chain and iteratively \emph{removes} the token whose deletion minimally degrades the pruning objective. 
The resulting deletion order induces a monotonic ranking over reasoning tokens, ordered by their contribution to preserving the pruning objective.
Unlike leave-one-out attribution~\cite{zintgraf2017visualizing}, which assigns independent importance scores under a fixed input, greedy pruning re-evaluates token importance under a changing context as pruning progresses, capturing token interactions (Section~\ref{subsec:dynamic_rankings}).
We illustrate greedy pruning in Figure~\ref{fig:main_figure} and formally describe it in Algorithm~\ref{alg:greedy_pruning}.
While greedy pruning is a myopic approximation and does not guarantee optimal subset selection, our claims focus on the existence and accessibility of internal ranking signals under likelihood-based probes, rather than the optimality of the induced subsets.
Greedy pruning is chosen for its simplicity; exploring richer search strategies, such as beam search, is left to future work.


\paragraph{Pruning Objectives.}
We consider two likelihood-based objectives that induce different notions of token importance with respect to answer generation. The first objective preserves answer likelihood,
\begin{equation}
\mathcal{L}^{\textsc{Ans}}_{\thetaP}(Q, R_K, A) = \log P_{\thetaP}(A \mid Q, R_K),
\end{equation}
allowing aggressive pruning of intermediate reasoning tokens provided that the pruner assigns high probability to the correct answer.
The second objective preserves both the reasoning and the answer under the model’s distribution,
\begin{equation}
\mathcal{L}^{\textsc{Joint}}_{\thetaP}(Q, R_K, A) = \log P_{\thetaP}(R_K, A \mid Q),
\end{equation}
and is therefore more restrictive, favoring pruned reasoning chains that remain consistent with the pruner’s reasoning trajectory while supporting answer generation. 
These objectives induce different notions of likelihood sensitivity over answers and reasoning structure, yielding distinct pruning behaviors under the model’s distribution (Section~\ref{subsec:functional_structure}).
All model likelihoods are computed conditioned on the gold token prefix; additional implementation details are provided in Appendix~\ref{app:train_and_inf}.

\paragraph{Interpretation of Pruning Ranks.}
Pruning ranks are not intended as faithful or causal explanations of a model’s internal computation, nor as human-interpretable reasoning steps.
Instead, they capture an interventional notion of likelihood sensitivity to token deletion under the model’s distribution.
Under this view, a token’s pruning rank reflects its functional importance for generation rather than mechanistic faithfulness.
To analyze patterns beyond tokenization artifacts, we aggregate tokens into higher-level semantic categories (Section~\ref{subsec:functional_structure}).
Overall, greedy pruning serves as a diagnostic probe for identifying reasoning-token subsequences that preserve predictive behavior.
We next instantiate greedy pruning in a teacher–pruner–student framework to evaluate its effect on distillation with compressed reasoning.

\begin{algorithm}[t]
\caption{Greedy Pruning}
\label{alg:greedy_pruning}
\begin{algorithmic}[1]
\Require Question $Q$; reasoning tokens $R = (r_1, \dots, r_n)$; answer $A$; objective $\mathcal{L}$; keep fraction $\rho \in (0,1]$; pruner model $\Pruner_{\theta_\Pruner}$
\Ensure Pruning ranks $\pi$; $K$, the set of token indices kept after pruning; pruned reasoning $R_K$, where $R_K$ denotes the subsequence of $R$ indexed by $K$, preserving the original order

\State $K \gets \{1,\dots,n\}$; \quad $m \gets \lceil \rho \cdot n \rceil$
\State Initialize $\pi[i] \gets \infty$ for all $i \in \{1,\dots,n\}$
\State $t \gets 1$
\While{$|K| > m$}
    \ForAll{$i \in K$}
        \State $L_i^{del} \gets \mathcal{L}_{\thetaP}(Q, R_{K \setminus \{i\}}, A)$
    \EndFor
    \State $i^\star \gets \arg\max_{i \in K} L_i^{del}$
    \State $\pi[i^\star] \gets t$
    \State $K \gets K \setminus \{i^\star\}$; \quad $t \gets t + 1$
\EndWhile \\
\Return $K, R_K, \pi$
\end{algorithmic}
\end{algorithm}


\section{Distillation Experiments}
\label{sec:experiments}

\begin{figure*}[t]
\centering


{\small
\legendmark{ub}{}{0pt}~\texttt{$\rho=1.0$}\quad
\legendmark{verbal}{star}{5pt}~\texttt{Greedy}\quad
\legendmark{meta}{isosceles triangle}{6pt}~\texttt{TokenSkip}\quad
\legendmark{entity}{rectangle}{5pt}~\texttt{H2O}\quad
\legendmark{symbolic}{diamond}{5pt}~\texttt{Uniform}\quad
\legendmark{grammar}{regular polygon}{5pt}~\texttt{Surprisal}\quad
\legendmark[dashed]{coref}{}{0pt}~\texttt{ZeroShot}
}

\vspace{0.5em}

\begin{subfigure}{0.24\textwidth}
  \centering
  \includegraphics[width=\linewidth]{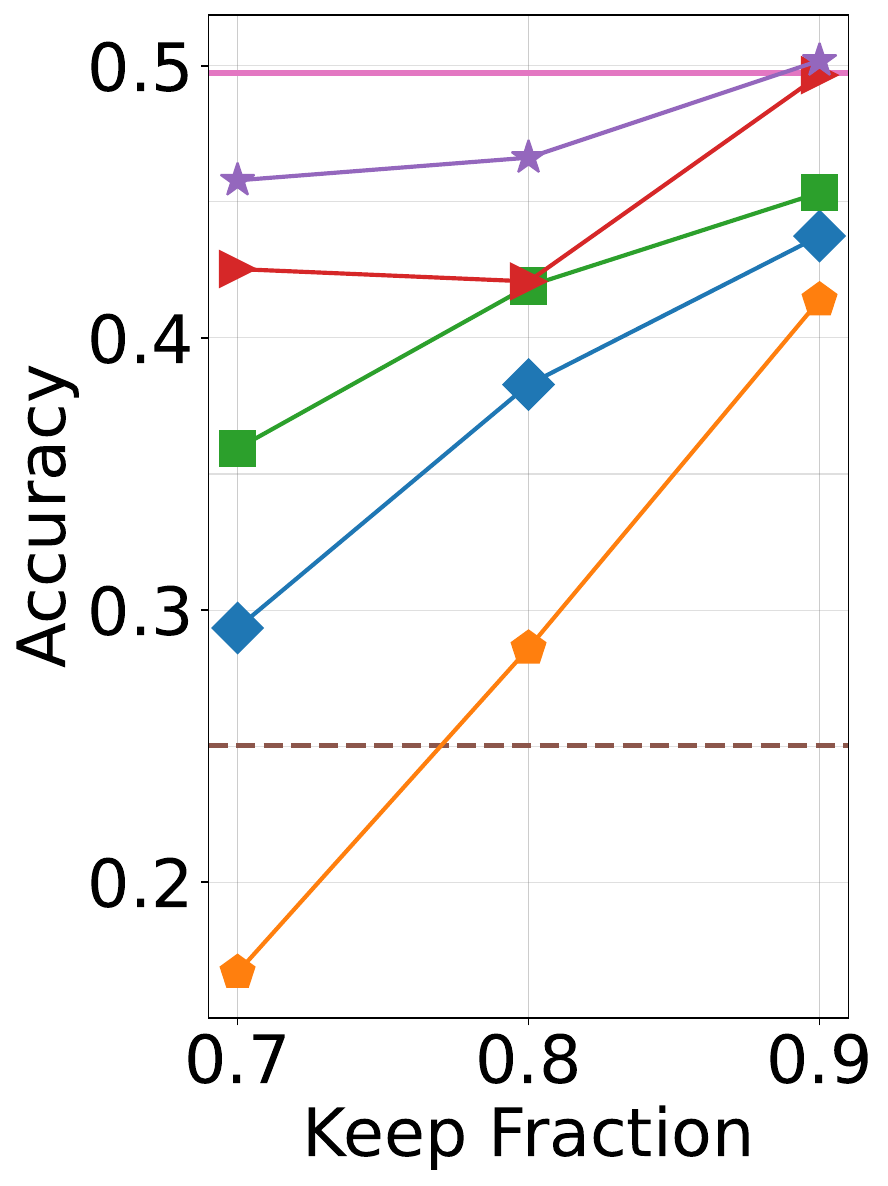}
\caption{{\fontsize{6.8pt}{6.8pt}\selectfont
$\Teacher,\Pruner=\text{Llama3.1-8B}$; GSM8K.}}
\label{fig:main-llama-2-gsm8k-llama3}
\end{subfigure}
\hfill
\begin{subfigure}{0.24\textwidth}
  \centering
  \includegraphics[width=\linewidth]{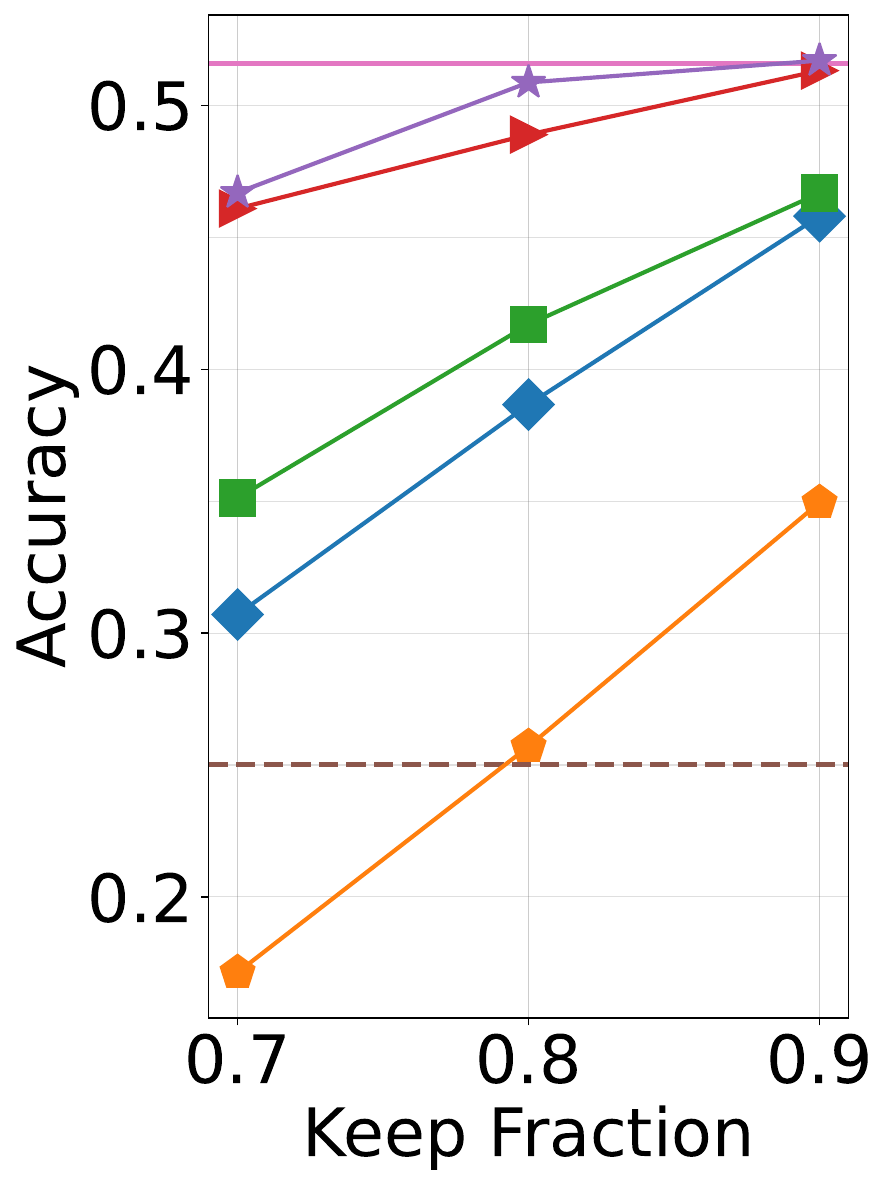}
\caption{{\fontsize{6.8pt}{6.8pt}\selectfont
$\Teacher,\Pruner=\text{Qwen2.5-7B}$; GSM8K.}}
\label{fig:main-llama-2-gsm8k-qwen2.5}
\end{subfigure}
\hfill
\begin{subfigure}{0.253\textwidth}
  \centering
  \includegraphics[width=\linewidth]{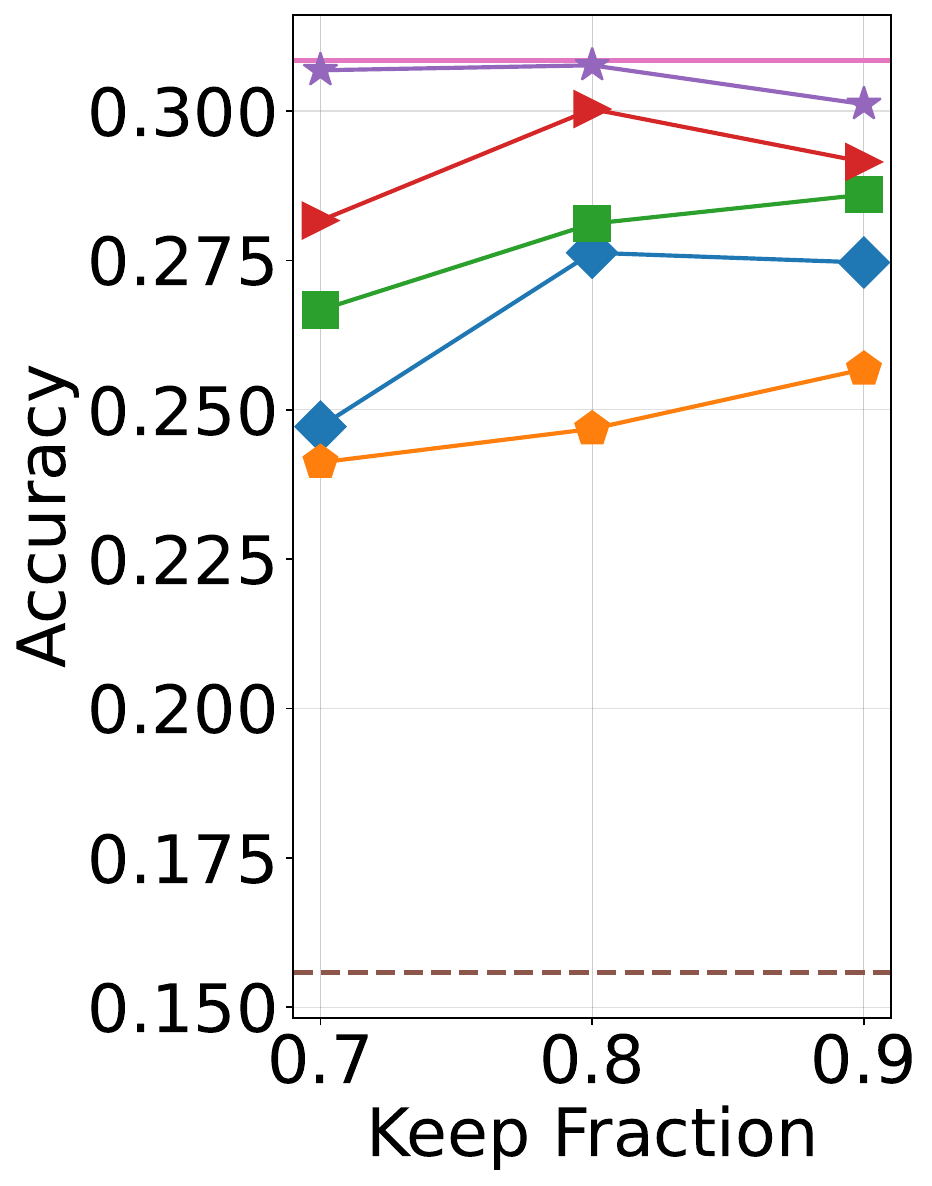}
\caption{{\fontsize{6.7pt}{6.7pt}\selectfont
$\Teacher,\Pruner=\text{Llama3.1-8B}$; MMLU-Pro.}}
\label{fig:main-llama-2-mmlu-llama-3}
\end{subfigure}
\hfill
\begin{subfigure}{0.248\textwidth}
  \centering
  \includegraphics[width=\linewidth]{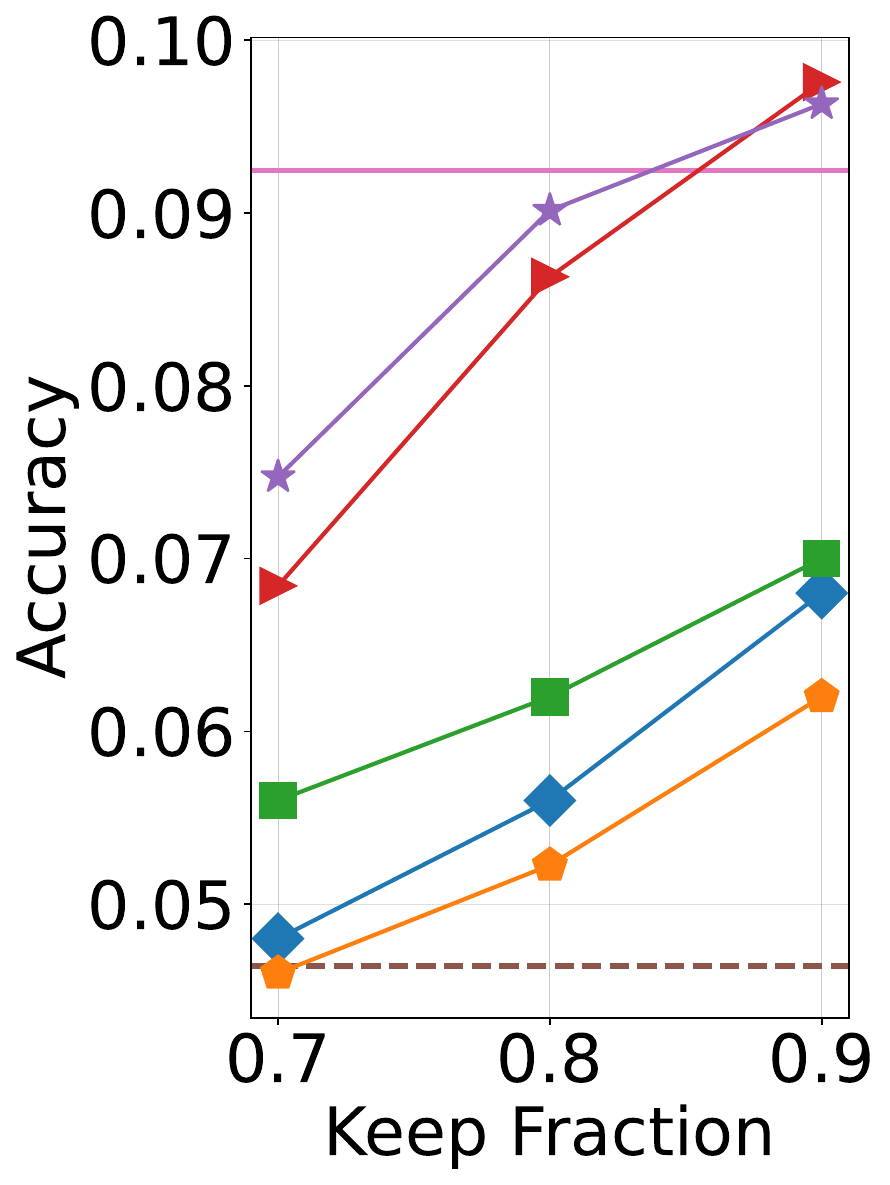}
\caption{{\fontsize{6.8pt}{6.8pt}\selectfont
$\Teacher,\Pruner=\text{Qwen2.5-7B}$; MATH.}}
\label{fig:main-llama-2-math500-qwen2.5}
\end{subfigure}

\caption{
\textbf{Distillation under reasoning token pruning.}
Accuracy of a Llama2-7B student trained on pruned reasoning at varying keep fractions, teacher, pruner, and dataset; dashed lines indicate zero-shot performance.
Greedy pruning achieves the strongest performance at matched lengths, indicating preservation of important tokens.
}



\label{fig:main-results-llama2}
\end{figure*}

\subsection{Experimental Setup}

We instantiate the teacher–pruner–student distillation framework (Section~\ref{sec:setup}) to evaluate greedy-pruned reasoning against multiple baselines.
If pruning preserves functionally important tokens for answer generation, the resulting reasoning chains should support effective student distillation.

\paragraph{Datasets and Models.}
We conduct experiments on three reasoning-intensive benchmarks: (i) GSM8K~\cite{Cobbe2021TrainingVT}, consisting of grade-school arithmetic word problems; (ii) MMLU-Pro~\cite{wang2024mmlupro}, a multi-domain benchmark spanning diverse subject areas; and (iii) MATH~\cite{hendrycksmath2021}, containing olympiad-level mathematical reasoning problems.
We use four instruction-tuned models spanning multiple families and performance levels: Llama-3.1-8B-Instruct, Qwen-2.5-7B-Instruct, Llama-2-7B-chat-hf, and Mistral-7B-Instruct, referred to as Llama3.1-8B, Qwen2.5-7B, Llama2-7B, and Mistral-7B, respectively.
Additional dataset and model details are provided in Appendix~\ref{app:model_datasets}.

\paragraph{Teacher--Pruner--Student Instantiation.}
We first evaluate all models in a zero-shot setting using deterministic greedy decoding (Table~\ref{tab:zero-shot}) and characterize relative performance per dataset.
Higher-performing models (Llama3.1-8B, Qwen2.5-7B) serve as teachers, while lower-performing models (Llama2-7B, Mistral-7B) serve as students, reflecting the intuition that stronger models induce more reliable reasoning chains and pruning signals that can be distilled into weaker models via compact reasoning supervision.
Reasoning chains are generated from the teacher using rejection sampling with temperature $0.7$.
The pruner then applies greedy pruning under a likelihood-based objective to produce compressed reasoning sequences at a specified keep fraction.
Unless otherwise stated, the pruner equals the teacher and the \textsc{Joint} objective is used.
Students are trained via supervised fine-tuning (SFT) on the pruned reasoning dataset and evaluated on downstream task accuracy.
Additional details are provided in Appendix~\ref{app:tps_config} and~\ref{app:train_and_inf}.

\paragraph{Baselines.}
To evaluate greedily pruned reasoning chains, we compare against four token-level importance baselines: \texttt{TokenSkip}~\cite{xia-etal-2025-tokenskip}, \texttt{H2O}~\cite{Zhang2023H2OHO}, \texttt{Surprisal}~\cite{li-etal-2023-compressing}, and \texttt{Uniform}. 
\texttt{TokenSkip} prunes reasoning tokens using a learned notion of semantic importance, with supervision from GPT-4~\cite{Achiam2023GPT4TR}. 
\texttt{H2O} is a cache-eviction method that ranks tokens by aggregated attention from future steps. 
\texttt{Surprisal} defines importance by token prediction probability, where higher probability implies lower surprisal and importance. 
\texttt{Uniform} is a non-informative baseline assigning equal importance to all tokens. 
For all comparisons, reasoning chains are identical and pruned to the same keep fraction, with token counts measured using the student tokenizer to account for tokenizer differences. 
Baseline details are provided in Appendix~\ref{app:baselines}.

\begin{figure*}[t]
\centering


\small
\legendmark{symbolic}{circle}{5pt}~\textsc{SymbMath}\quad
\legendmark{grammar}{star}{5pt}~\textsc{Grammar}\quad
\legendmark{entity}{isosceles triangle}{6pt}~\textsc{EntName}\quad
\legendmark{meta}{rectangle}{5pt}~\textsc{MetaDisc}\quad
\legendmark{verbal}{regular polygon}{5pt}~\textsc{VerbalMath}\quad
\legendmark{coref}{diamond}{5pt}~\textsc{CoRef}

\vspace{0.5em}

\begin{subfigure}{0.31\textwidth}
  \centering
  \includegraphics[width=\linewidth]{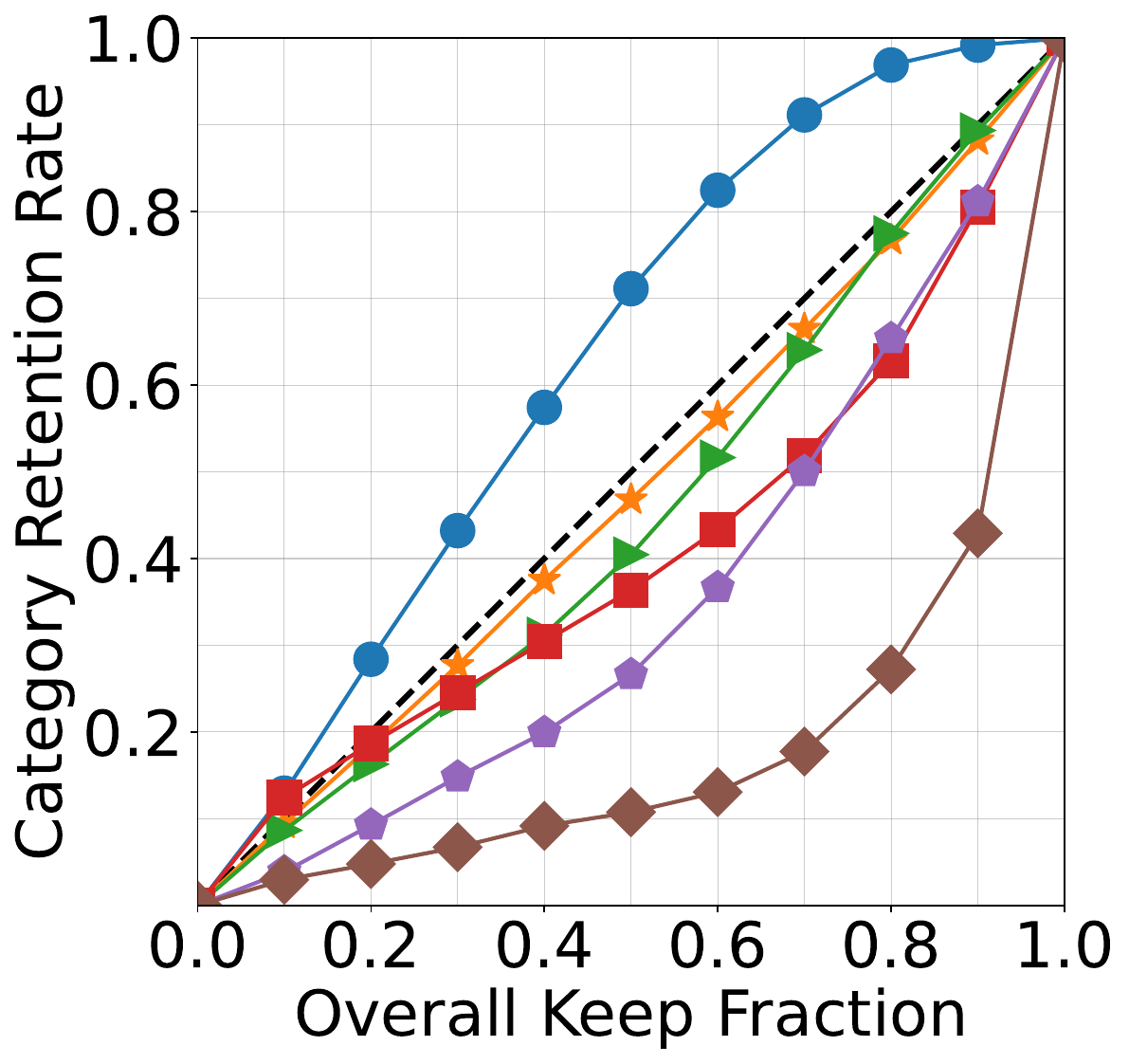}
\caption{{\fontsize{7.0pt}{7.0pt}\selectfont
$\Teacher,\Pruner=\text{Qwen2.5-7B};\ \mathcal{L}^{\textsc{Joint}}_{\theta_\Pruner}$.}}
\label{fig:functional-structure-a}
\end{subfigure}
\hfill
\begin{subfigure}{0.31\textwidth}
  \centering
  \includegraphics[width=\linewidth]{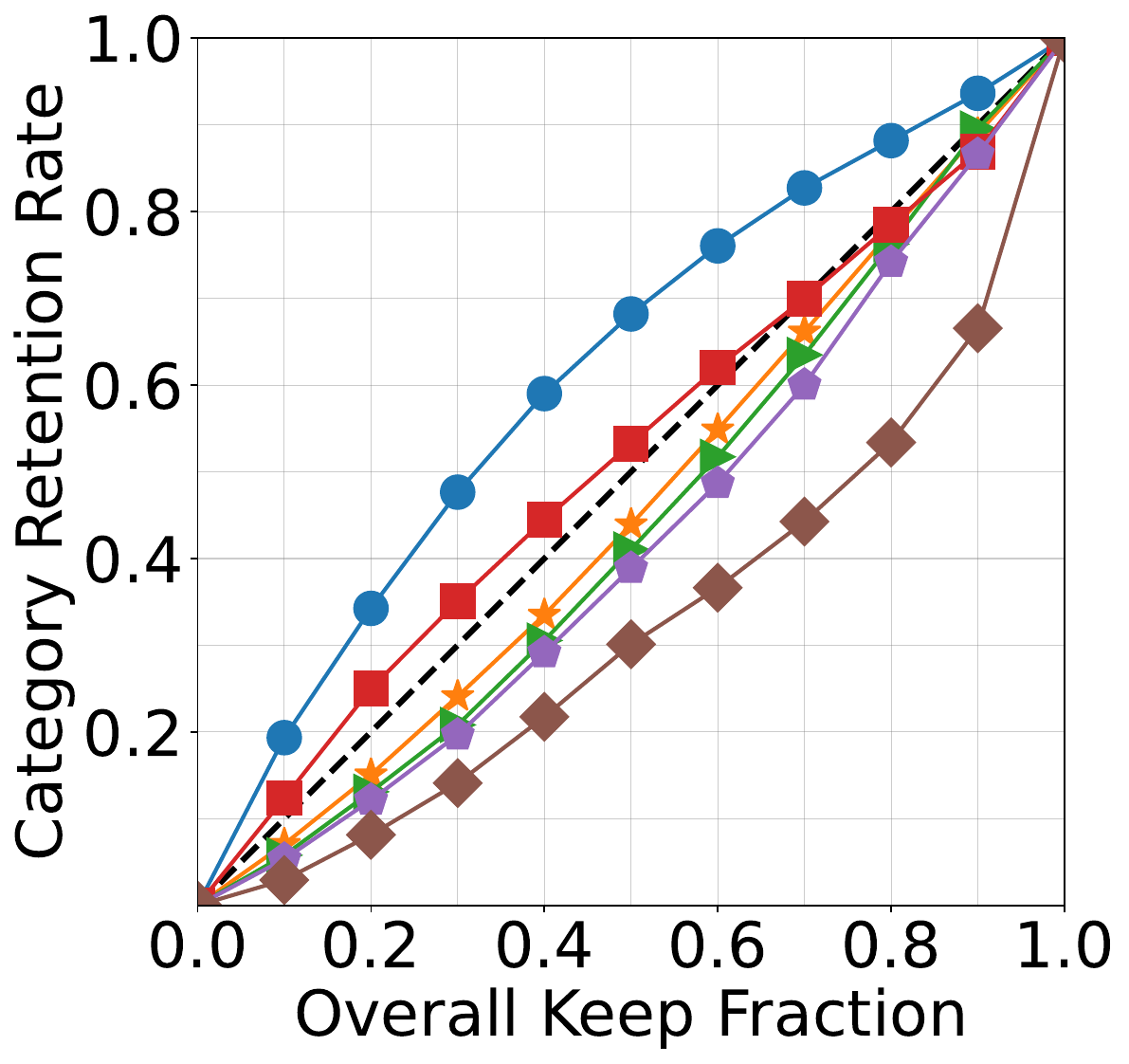}
\caption{{\fontsize{7.0pt}{7.0pt}\selectfont
$\Teacher,\Pruner=\text{Qwen2.5-7B};\ \mathcal{L}^{\textsc{Ans}}_{\theta_\Pruner}$.}}
\label{fig:functional-structure-b}
\end{subfigure}
\hfill
\begin{subfigure}{0.31\textwidth}
  \centering
  \includegraphics[width=\linewidth]{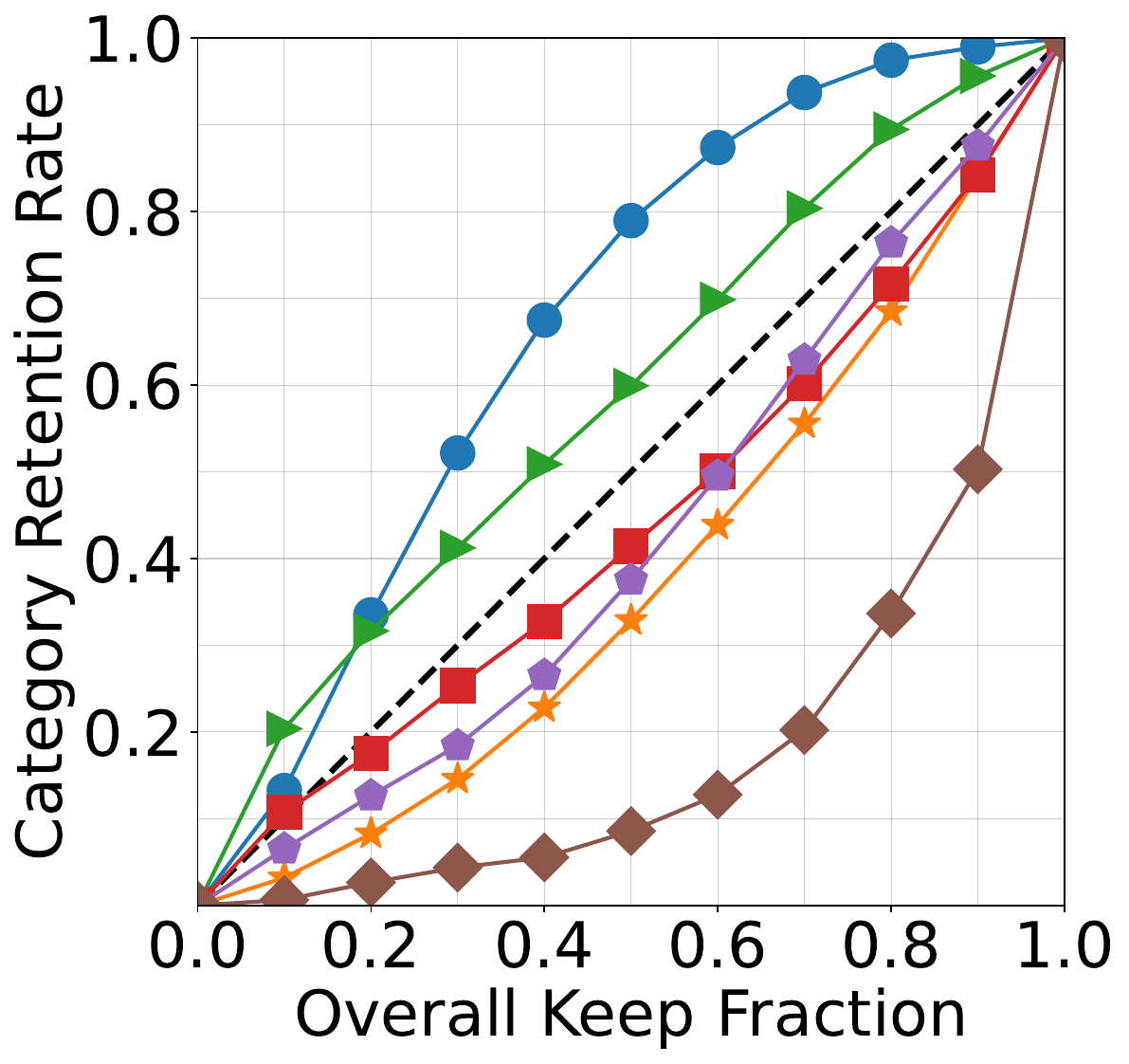}
\caption{{\fontsize{6.8pt}{6.8pt}\selectfont
$\Teacher=\text{Qwen2.5-7B};\ \Pruner=\text{Llama2-7B};\ \mathcal{L}^{\textsc{Joint}}_{\theta_\Pruner}$.}}
\label{fig:functional-structure-c}
\end{subfigure}

\caption{{
\textbf{Functional structure under greedy pruning.}
Each curve shows the fraction of tokens retained per category at a given keep fraction. Panels vary teacher, pruner, and pruning objective; the dashed line indicates uniform pruning.
(a) Pruning preferentially preserves symbolic computation while removing referential, descriptive, and linguistic scaffolding.
(b) Excluding reasoning likelihood in pruning objective softens induced functional structure.
(c) A weaker pruner preserves symbolic computation but disrupts the balance of non-symbolic structure.
}}

\label{fig:functional-structure-main}
\end{figure*}

\subsection{Results}
\label{subsec:distill-results}

Figure~\ref{fig:main-results-llama2} reports downstream accuracy of student models trained on pruned reasoning traces across keep fractions, teacher–pruner configurations, and datasets. All pruning methods operate on identical teacher-generated reasoning chains, and all students are trained with the same SFT protocol. Under this controlled setup, distillation performance tests whether pruned reasoning retains the information needed for correct answer generation.

Across all settings, greedy pruning yields the strongest student performance among pruning-based methods at matched keep fractions. Accuracy improves smoothly as the keep fraction increases, indicating graceful degradation under compression. On GSM8K, students trained on greedy-pruned reasoning outperform \texttt{TokenSkip}, \texttt{H2O}, \texttt{Surprisal}, and \texttt{Uniform} across all keep fractions, for both Llama3.1-8B and Qwen2.5-7B teacher–pruner pairs (Figures~\ref{fig:main-llama-2-gsm8k-llama3}, \ref{fig:main-llama-2-gsm8k-qwen2.5}). \texttt{Uniform} and \texttt{Surprisal} drop sharply at aggressive pruning, while \texttt{TokenSkip} improves with higher keep fractions but remains below greedy pruning. Zero-shot student performance is shown for reference and is consistently worse than all pruning-based distillation methods. Similar trends hold on the more challenging MMLU-Pro and MATH benchmarks (Figures~\ref{fig:main-llama-2-mmlu-llama-3}, \ref{fig:main-llama-2-math500-qwen2.5}): despite lower absolute accuracy, greedy pruning remains best at matched reasoning lengths. Additional results with Mistral-7B as the student show similar trends (Appendix~\ref{sec:add_dist}). 
Overall, pruned reasoning supports effective distillation across datasets, models, and keep fractions, consistently outperforming the frontier-model–supervised \texttt{TokenSkip} baseline.

We further analyze three key aspects of our compressed-reasoning distillation approach.
First, we quantify the computational cost of greedy pruning (Appendix~\ref{app:compute_cost}), showing that while the naive implementation scales superlinearly with sequence length, it is a one-time offline preprocessing step and admits practical optimizations. Second, we evaluate the readability and structural integrity of pruned traces (Appendix~\ref{app:corruption}), finding that even at 30\% token reduction, degradation is primarily surface-level, with semantic coherence and mathematical state largely preserved. Third, we study the behavior of students trained on pruned data (Appendix~\ref{app:student_behavior}), observing that they learn to generate proportionally shorter reasoning traces without meaningful loss in semantics or structure.

Together, these results indicate that greedy pruning removes redundant tokens while preserving those critical for answer generation, motivating a closer analysis of the induced importance structure, which we study next.

\section{Analysis of Pruning Behavior}
\label{sec:analysis}

\subsection{Functional Structure Under Pruning}
\label{subsec:functional_structure}

To analyze which reasoning tokens are preferentially preserved or removed under greedy pruning, we annotate each reasoning token by its \emph{functional role in reasoning}, complementing prior reasoning-step–level analyses of reasoning structure~\cite{Marjanovi2025DeepSeekR1TL} with a token-level functional characterization. This annotation allows us to track how different functional components evolve across pruning stages and to characterize the emergent structure of pruned reasoning chains. We conduct this analysis on 1{,}000 randomly sampled GSM8K examples, as GSM8K offers particularly interpretable token-level functional categories; extending this analysis to additional datasets is left for future work. Specifically, we define six coarse, interpretable functional categories: \textsc{SymbMath} (explicit equations and mathematical symbols), \textsc{MetaDisc} (reasoning narration and structuring), \textsc{CoRef} (pronouns and referential expressions), \textsc{EntName} (concrete entities), \textsc{VerbalMath} (natural-language arithmetic relations), and \textsc{Grammar} (grammatical fillers). Annotation is performed using \texttt{gpt-5-mini} and validated through manual inspection and stability checks; details are provided in Appendix~\ref{sec:appendix-categorization-scheme} and~\ref{subsec:annotation_process}.

\paragraph{Core functional structure.}
Figures~\ref{fig:functional-structure-a} and~\ref{fig:functional-structure-app-a} (Appendix~\ref{sec:appendix-additional-plots}) plot the \emph{category retention rate} as a function of the \emph{overall keep fraction} for teacher–pruner settings where both models are Qwen2.5-7B and Llama3.1-8B, respectively, under the \textsc{Joint} objective. Each point at keep fraction $k$ reports the average fraction of tokens within a category that remain after pruning to $k$, with the dashed diagonal indicating a uniform deletion baseline. Across both settings, greedy pruning exhibits a stable and interpretable functional ordering. \textsc{SymbMath} tokens are strongly over-retained throughout pruning, deviating substantially above the uniform baseline, indicating that explicit symbolic computations are preserved disproportionately even under aggressive compression. In contrast, \textsc{CoRef} tokens are markedly under-retained until late pruning stages, suggesting that explicit referential bookkeeping is among the earliest information removed. \textsc{VerbalMath} tokens also exhibit systematically lower retention than \textsc{SymbMath}, falling below the diagonal across a wide range of keep fractions, while \textsc{Grammar}, \textsc{MetaDisc}, and \textsc{EntName} tokens lie closer to the uniform baseline. 
Together with the token-frequency plots in Appendix~\ref{sec:appendix-additional-plots}, these findings show that greedy pruning goes beyond category frequency, selectively preserving symbolic computation while removing referential, descriptive, and linguistic scaffolding.
We provide analogous functional-structure analyses for baseline methods in Appendix~\ref{sec:appendix-additional-plots}.

\begin{figure}[t]
\centering

\small
\legendmark{verbal}{star}{5pt}~\texttt{Base}\quad
\legendmark{coref}{circle}{5pt}~$\mathcal{L}^{\textsc{Ans}}_{\thetaP}$ \quad
\legendmark{symbolic}{isosceles triangle}{6pt}~$\Pruner=$ Llama2-7B

\vspace{0.5em}

\includegraphics[width=0.55\linewidth]{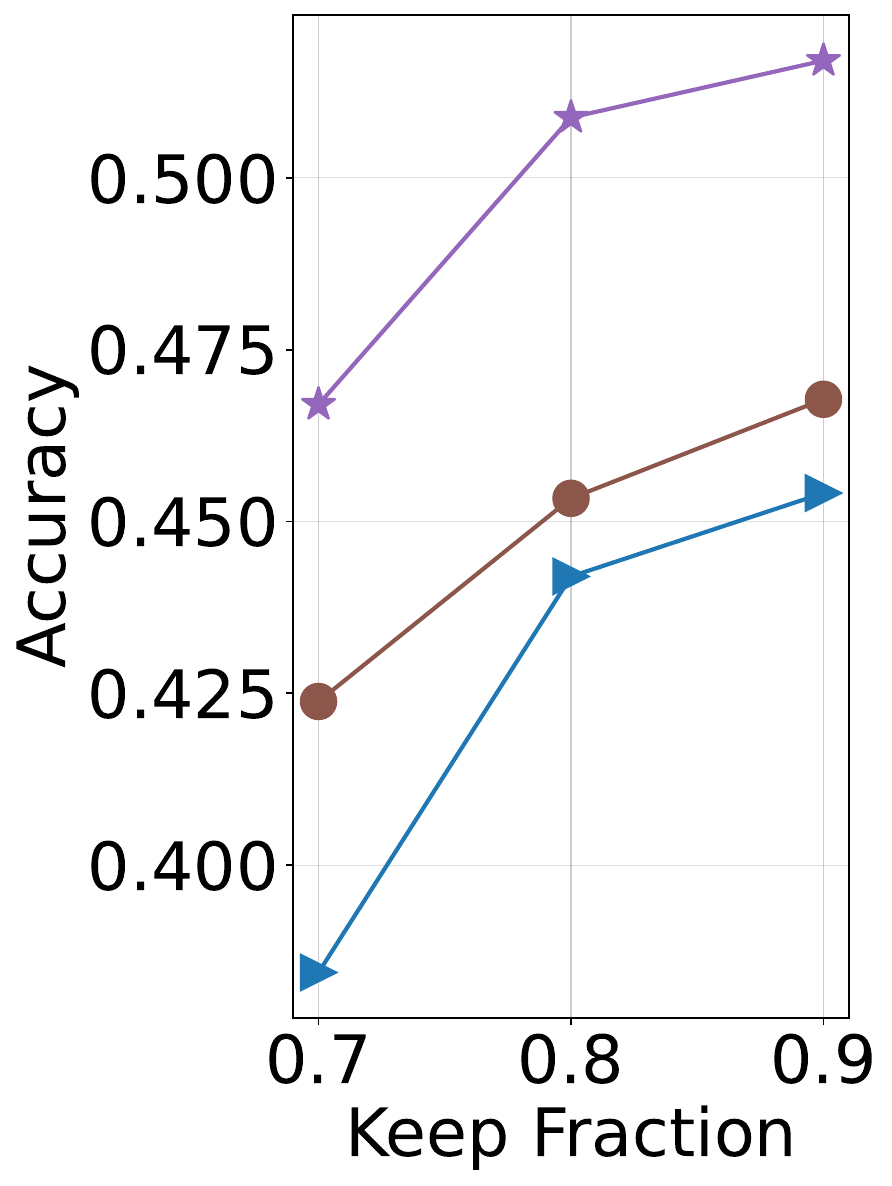}

\caption{
\textbf{Effect of pruning objective and pruner strength on distillation.}
The base greedy setting uses $\Teacher,\Pruner=\text{Qwen2.5-7B}$ and $\Student=\text{Llama2-7B}$ with the \textsc{Joint} objective.
We ablate pruner strength ($\Pruner=\text{Llama2-7B}$) and pruning objective (answer-only, \textsc{Ans}).
Both weaken student performance, with pruner strength having the larger effect.
}

\label{fig:ablation}
\end{figure}

\paragraph{Effect of pruning objective.}
We next analyze how the pruning objective influences the induced functional structure and its relationship to downstream distillation performance. Fixing a single teacher–pruner–student configuration ($\Teacher,\Pruner=\text{Qwen2.5-7B}$, $\Student=\text{Llama2-7B}$) on GSM8K, we compare the \textsc{Joint} objective to an answer-only (\textsc{Ans}) objective. As shown in Figure~\ref{fig:functional-structure-b}, answer-only pruning preserves coarse functional ordering, with \textsc{SymbMath} remaining over-retained and \textsc{CoRef} pruned early. However, separation across functional categories is weakened: retention curves for \textsc{VerbalMath}, \textsc{MetaDisc}, and \textsc{EntName} largely collapse toward the uniform baseline, and even \textsc{SymbMath} and \textsc{CoRef} move closer to uniform deletion. This attenuation of functional structure aligns with lower student accuracy observed under the \textsc{Ans} objective (Figure~\ref{fig:ablation}), indicating that incorporating reasoning likelihood sharpens the structure revealed by greedy pruning and yields reasoning chains more effective for distillation.

\paragraph{Effect of pruner model strength.}
Finally, we study how pruner model strength affects the functional structure induced by greedy pruning and its downstream impact. Using the same configuration as for pruning objective, we replace the Qwen2.5-7B pruner with a weaker Llama2-7B pruner. Comparing Figures~\ref{fig:functional-structure-a} and~\ref{fig:functional-structure-c}, the overall ordering of \textsc{SymbMath} and \textsc{CoRef} tokens remains stable, with symbolic computation preferentially retained and coreference pruned in both settings. In contrast, the treatment of non-symbolic categories shifts: \textsc{EntName} tokens move from being under-retained to over-retained, \textsc{Grammar} tokens are pruned more aggressively, and \textsc{VerbalMath} moves closer to the uniform deletion baseline. This altered functional structure coincides with a larger drop in student accuracy observed under weaker pruner (Figure~\ref{fig:ablation}), suggesting that effective distillation depends not only on preserving symbolic computation but also on maintaining an appropriate balance of non-symbolic structure.

\begin{figure}[t]
\centering
\includegraphics[width=0.72\linewidth]{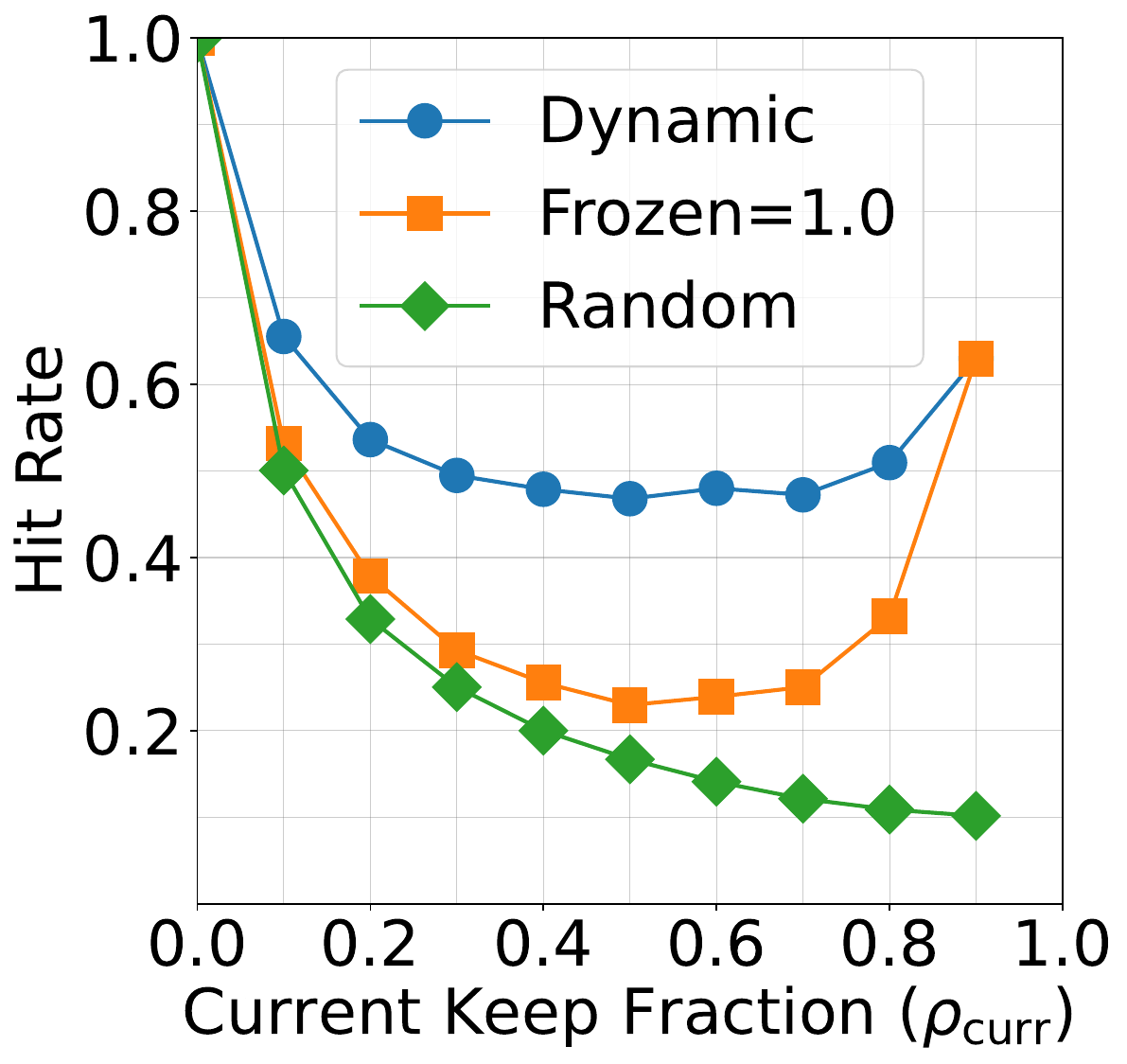}

\caption{
\textbf{Dynamics of pruning ranks.}
\texttt{Hit@|S|} alignment between tokens removed at keep fraction $\rho_{\text{curr}}$ and $L^{del}$-based local ranks at the previous pruning stage.
Dynamic rankings ($\rho_{\text{prev}}=\rho_{\text{curr}}+0.1$) consistently outperform frozen ($\rho_{\text{prev}}{=}1.0$) and random baselines across keep fractions, indicating that greedy pruning re-evaluates token importance as context contracts.
}
\label{fig:dynamic_rankings}
\end{figure}

\subsection{Dynamics of Pruning Ranks}
\label{subsec:dynamic_rankings}

A natural question raised by greedy pruning is whether it reveals a fixed global ordering of token importance or whether importance is dynamically reshaped as pruning progresses. If importance were static, rankings computed early in pruning should remain predictive. Here, we test whether greedy pruning instead continually re-evaluates token importance as the retained context contracts, a property that could help explain its advantage over baselines relying on static importance signals.

Greedy pruning (Algorithm~\ref{alg:greedy_pruning}) assigns pruning ranks sequentially using post-deletion likelihoods as importance signals. At keep fraction $\rho$, for the remaining token set $K_\rho$, the pruner evaluates
\[
L_i^{\text{del}} = \mathcal{L}_{\theta_{\Pruner}}(Q, R_{K_\rho \setminus \{i\}}, A)
\]
for each $i \in K_\rho$. Tokens with higher $L_i^{\text{del}}$ incur smaller likelihood drops and are considered safer to remove earlier, inducing a \emph{local pruning ranking} at each pruning stage. All experiments use the same GSM8K subset as Section~\ref{subsec:functional_structure}, with Qwen2.5-7B as both teacher and pruner under the \textsc{Joint} objective.

Figure~\ref{fig:dynamic_rankings} measures how well local pruning rankings predict subsequent pruning decisions across keep fractions. At each transition from $\rho_{\text{prev}}$ to $\rho_{\text{curr}}$, a set of tokens
\[
S_{\rho_{\text{curr}}} = K_{\rho_{\text{prev}}} \setminus K_{\rho_{\text{curr}}}
\]
is pruned. We evaluate alignment using \texttt{Hit@|S|}, defined as the overlap between the top-ranked tokens under the local ranking at $\rho_{\text{prev}}$ and the tokens actually pruned at $\rho_{\text{curr}}$.

Dynamic local rankings consistently outperform both a \texttt{Frozen=1.0} baseline, which reuses rankings computed at $\rho=1.0$ to simulate a fixed global ordering, and a random baseline. The widening gap between dynamic and frozen rankings, especially at intermediate keep fractions, shows that greedy pruning does not expose a static importance ordering early in pruning. Instead, token importance is continually re-evaluated as the reasoning chain contracts, with later-stage local rankings remaining most predictive of imminent pruning decisions. This dynamic behavior highlights greedy pruning as an adaptive process rather than one governed by a fixed global notion of importance.

\subsection{Predicting Pruning Ranks from Attention}
\label{subsec:attention-surrogate}

We test whether the pruning ranks induced by greedy pruning are predictable from attention patterns alone. Inspired by prior work on attribution~\cite{CohenWang2025LearningTA}, we train a two-layer MLP surrogate to predict token-level pruning scores using only attention signals from the teacher model. 
For each reasoning token, we construct a feature vector of dimension $[\texttt{\#layers} \times \texttt{\#heads}]$ by aggregating the average attention it receives from subsequent tokens across all future positions at each attention head.

The surrogate is trained to predict the post-deletion likelihood $L^{del}$ associated with removing each token under greedy pruning, using a Pearson correlation objective that emphasizes relative importance rather than absolute scale. This setup allows us to test whether attention patterns encode sufficient information to predict the pruning order of reasoning tokens. Using only 200 GSM8K examples for training with Llama3.1-8B as the teacher, the surrogate achieves a Pearson correlation of 0.88 with true post-deletion likelihoods on a held-out set of 1,000 GSM8K examples. These results indicate that attention patterns encode information strongly predictive of greedy pruning decisions and suggest a promising direction for incorporating such signals into more efficient pruning or search procedures.

\section{Discussion \& Future Directions}

Beyond compression, greedy pruning also serves as a diagnostic probe of model-internal reasoning. By selectively removing tokens while preserving output likelihood, it reveals which parts of a model's generated reasoning are functionally necessary for producing correct answers. Our results (\Cref{subsec:functional_structure}) further show that pruning behavior depends on both the pruning objective and the pruner's capacity, suggesting that the induced importance structure reflects the properties of the underlying model. Understanding how this structure varies across models, scales, and training procedures opens a promising avenue for future work.

Beyond these diagnostic implications, greedy pruning can be viewed as a structured deletion operator for compressing reasoning chains under likelihood-based objectives, offering a controlled mechanism for exploring reasoning space in LLMs. From this perspective, it represents one instance of a broader class of reasoning-space operators. Extending this framework beyond deletion to operations such as token insertion or replacement under alternative objectives could enable finer-grained manipulation of reasoning trajectories, providing a more principled alternative to purely sampling-based control.

The tractability of such extensions is further supported by our finding that pruning ranks are strongly predictable from attention-based signals, suggesting that token-level importance may be partially accessible during generation. This connects naturally to recent work on inference-time pruning~\cite{yang2025pencil,Monea2025BreadcrumbsRM}, which discards completed reasoning segments to manage context length. While our work focuses on likelihood-preserving post-hoc pruning as an analysis and distillation tool, extending similar objectives to inference-time settings or incorporating pruning signals into training-time curricula represents a natural next step toward more efficient and controllable reasoning systems.

Together, these directions suggest that pruning-based objectives may serve as a unifying interface for analyzing and shaping reasoning in LLMs, not merely as a compression tool but as a principled lens into model cognition.




\section{Conclusion}

We investigate whether LLMs encode token-level functional importance for answer generation within their reasoning chains.
Using greedy pruning, a likelihood-preserving deletion procedure, we show that models can systematically compress reasoning while retaining information critical for answer generation, and that students trained on such pruned reasoning outperform multiple baselines at matched reasoning lengths. Our analyses reveal a stable and interpretable functional structure under pruning, with symbolic computation preferentially preserved and supporting linguistic scaffolding pruned earlier. Furthermore, we show that pruning behavior depends on the pruning objective and pruner strength. We also demonstrate that pruning ranks are dynamically reshaped as context contracts and are strongly predictable from attention patterns, indicating that models encode internal signals of reasoning-token importance. Together, these findings establish greedy pruning as a diagnostic method for exposing token-level importance structure in model-generated reasoning.

\section*{Limitations}

We study likelihood-based greedy pruning as a tool for exposing token-level importance structure in LLM reasoning. A primary limitation is computational cost: greedy pruning requires repeated likelihood evaluations over candidate token deletions, leading to superlinear scaling in sequence length and making naive implementations expensive for long reasoning chains. In our setting, this cost is incurred as a one-time offline preprocessing step, amortized over downstream student training. Importantly, our goal is not to provide a universally scalable pruning algorithm but to analyze token-level importance at realistic reasoning lengths (e.g., $\leq$500 tokens), where the method remains empirically tractable.

Several directions could reduce this cost. First, prefix caching\footnote{\url{https://docs.vllm.ai/en/stable/design/prefix_caching/}} can reuse KV states for shared prefixes across deletion candidates, yielding constant-factor speedups. Second, the stability of local pruning ranks (\Cref{subsec:dynamic_rankings}) suggests that multiple tokens could be removed per iteration, reducing the number of recomputation steps. Third, attention-based or learned surrogate models could approximate token importance without full likelihood recomputation, supported by the strong correlation between post-deletion likelihoods and attention-derived features (\Cref{subsec:attention-surrogate}). A detailed analysis of computational costs and scaling behavior is provided in \Cref{app:compute_cost}.

Beyond efficiency, aggressive pruning can degrade the readability of reasoning traces. While moderate pruning primarily introduces formatting-level artifacts rather than loss of semantic or mathematical structure (\Cref{app:corruption}), we do not characterize the precise threshold at which reasoning becomes unsuitable for human interpretation. Developing principled criteria for reasoning interpretability under compression remains an open problem.

Our analysis is further restricted to teacher-generated, correctness-filtered reasoning traces. How pruning behaves on incorrect or low-quality reasoning, where likelihood signals may be less aligned with functional importance, remains an open question.

Our experiments are conducted on moderately sized instruction-tuned LLMs and do not extend to very large reasoning-focused models (e.g., DeepSeek-style~\cite{guo2025deepseek}) due to computational constraints. That said, our ablations indicate that pruning quality is strongly governed by pruner strength (\Cref{subsec:functional_structure}), suggesting that larger models may yield more reliable token-importance estimates and that the qualitative trends observed here are likely to persist or strengthen at scale.

\section*{Acknowledgments}
This work used the DeltaAI system at the National Center for Supercomputing Applications [award OAC 2320345] through allocation CIS251207 from the Advanced Cyberinfrastructure Coordination Ecosystem: Services \& Support (ACCESS) program, which is supported by National Science Foundation grants \#2138259, \#2138286, \#2138307, \#2137603, and \#2138296.

\newpage

\bibliography{acl}

\appendix
\renewcommand{\sectionautorefname}{Appendix}

\newpage

\section{LLM Usage}
Other than being used as part of the experiments conducted in this work, LLMs were used solely as a writing assistance tool in preparing this paper submission. Their role was limited to polishing language, improving clarity, and reducing redundancy. The prompt used for this purpose was similar to ``Please revise the writing of this, making sure to remove any grammatical mistakes.'' All research ideas, experimental designs, analyses, and claims presented in the paper are entirely the original work of the authors. No part of the conceptual, methodological, or empirical contributions relies on or originates from LLM outputs.

\newpage

\section{Implementation Details}
\label{sec:implementation_details}

\subsection{Datasets}
\label{app:model_datasets}

We conduct experiments on three reasoning-intensive benchmarks:
(i) GSM8K~\cite{Cobbe2021TrainingVT}, consisting of grade-school arithmetic word problems;
(ii) MMLU-Pro~\cite{wang2024mmlupro}, a multi-domain benchmark spanning diverse domains or subjects; and
(iii) MATH~\cite{hendrycksmath2021}, containing Olympiad-level mathematical reasoning problems.
Hugging Face identifiers for original datasets used are summarized in Table~\ref{table:model_details}.
As these benchmarks differ in their provided train/dev/test splits, we standardize the evaluation protocol by constructing explicit training, development, and test splits as described below.

\begin{itemize}[itemsep=1pt, topsep=5pt, leftmargin=*]
  \item \textbf{GSM8K.}
The original dataset provides 7.47K training examples and 1.32K test examples.
We randomly partition the training split into 6.97K training examples and 0.5K development examples while keeping the original test split intact.
The resulting split is released at \href{https://huggingface.co/datasets/iamjanvijay/gsm8k}{\texttt{iamjanvijay/gsm8k}}.

\item \textbf{MMLU-Pro.}
The original dataset does not provide a standard training split.
We therefore randomly partition the available 12K examples into 10K training, 0.8K development, and 1.23K test examples.
The resulting split is released at \href{https://huggingface.co/datasets/iamjanvijay/MMLU-Pro}{\texttt{iamjanvijay/MMLU-Pro}}.

\item \textbf{MATH.}
The dataset provides 12K training examples and a 500-example test set.
We randomly split the training data into 11.2K training examples and 0.8K development examples and evaluate on the original MATH-500 test set.
The resulting split is released at \href{https://huggingface.co/datasets/iamjanvijay/openaimath}{\texttt{iamjanvijay/openaimath}}.
\end{itemize}

We use these fixed, constructed splits to conduct all the experiments.

\begin{table}[h]
\centering
\small
\setlength{\tabcolsep}{4pt}
\renewcommand{\arraystretch}{1.0}
\begin{tabular}{ll}
\toprule
\textbf{Shorthand} & \textbf{HuggingFace Identifier} \\
\midrule
Llama3.1-8B     & \href{https://huggingface.co/meta-llama/Llama-3.1-8B-Instruct}{\texttt{meta-llama/Llama-3.1-8B-Instruct}} \\
Qwen2.5-7B     & \href{https://huggingface.co/Qwen/Qwen2.5-7B-Instruct}{\texttt{Qwen/Qwen2.5-7B-Instruct}} \\
Llama2-7B     & \href{https://huggingface.co/meta-llama/Llama-2-7b-chat-hf}{\texttt{meta-llama/Llama-2-7b-chat-hf}} \\
Mistral-7B     & \href{https://huggingface.co/mistralai/Mistral-7B-Instruct-v0.3}{\texttt{mistralai/Mistral-7B-Instruct-v0.3}} \\
\midrule
GSM8K & \href{https://huggingface.co/datasets/openai/gsm8k}{\texttt{openai/gsm8k}} \\
MMLU-Pro    & \href{https://huggingface.co/datasets/TIGER-Lab/MMLU-Pro}{\texttt{TIGER-Lab/MMLU-Pro}} \\
MATH  & \href{https://huggingface.co/datasets/simplescaling/openaimath}{\texttt{simplescaling/openaimath}} \\

\bottomrule
\end{tabular}
\caption{Mapping from shorthand model and dataset names to their corresponding Hugging Face identifiers.}
\label{table:model_details}
\end{table}

\subsection{Teacher--Pruner--Student Configuration}
\label{app:tps_config}

We instantiate the teacher--pruner--student distillation framework with four instruction-tuned models and the datasets listed in Table~\ref{table:model_details}.
To justify model role assignments, we first evaluate each model’s zero-shot performance on the corresponding test splits (Table~\ref{tab:zero-shot}).
All zero-shot evaluations use deterministic greedy decoding (\texttt{temperature=0.0, n=1, top\_p=1.0}).
Based on these results, we designate Qwen2.5-7B and Llama3.1-8B as teacher models and Llama2-7B and Mistral-7B as student models.
This choice follows standard distillation practice: stronger teachers provide higher-quality reasoning supervision, while weaker students ensure meaningful headroom for knowledge and reasoning transfer.
Unless otherwise stated, we set the pruner model equal to the teacher to avoid confounding pruning quality with pruner capacity.
To isolate the effect of pruner strength (Section~\ref{subsec:functional_structure}), we additionally instantiate the pruner as Llama2-7B while holding the teacher fixed.
We use a shared prompt per dataset to ensure comparability across experiments; the exact system prompts and user prompt templates are provided below.

\begin{table}[t]
\centering
\small
\setlength{\tabcolsep}{6pt}
\renewcommand{\arraystretch}{0.8}
\begin{tabular}{lccc}
\toprule
\textbf{Model} & \textbf{GSM8K} & \textbf{MMLU-Pro} & \textbf{MATH-500} \\
\midrule
Llama2-7B     & 25.02 & 15.58 &  4.60 \\
Mistral-7B   & 31.46 & 30.28 & 13.40 \\
Qwen2.5-7B   & 90.98 & 56.25 & 65.80 \\
Llama3.1-8B  & 82.34 & 45.21 & 43.60 \\
\bottomrule
\end{tabular}
\caption{\textbf{Zero-shot Performance.} Accuracy (\%) of models on the test splits of the datasets.}
\label{tab:zero-shot}
\end{table}

\newpage

\begin{tcolorbox}[
  breakable,
  colback=white,
  colframe=black,
  title={Prompt: GSM8K},
  label={temp},
  title after break={Prompt: GSM8K Teacher Model (continued)},
]
\label{box:annotation-prompt}

\setlength{\parindent}{0pt}
\renewcommand{\baselinestretch}{0.92}\selectfont

\begin{lstlisting}

(*@\textbf{SYSTEM PROMPT}@*)

You are a helpful assistant. Please solve the following problem step by step. At the end, output the final **answer only in the JSON format**:\n\n{"answer": "[numeric answer only]"}

(*@\textbf{USER PROMPT TEMPLATE}@*)

{{question}}

\end{lstlisting}
\end{tcolorbox}

\begin{tcolorbox}[
  breakable,
  colback=white,
  colframe=black,
  title={Prompt: MMLU-Pro},
  label={temp},
  title after break={Prompt: MMLU-Pro Teacher Model (continued)},
]
\label{box:annotation-prompt}

\setlength{\parindent}{0pt}
\renewcommand{\baselinestretch}{0.92}\selectfont

\begin{lstlisting}

(*@\textbf{SYSTEM PROMPT}@*)

You are a helpful assistant. Please solve the following MCQ problem step by step to select the correct answer from the given options. At the end, output the final answer only in the following format: "Thus, the final answer is: [correct letter choice only]"

(*@\textbf{USER PROMPT TEMPLATE}@*)

{{question}}

\end{lstlisting}
\end{tcolorbox}

\begin{tcolorbox}[
  breakable,
  colback=white,
  colframe=black,
  title={Prompt: MATH-500},
  label={temp},
  title after break={Prompt: MATH-500 Teacher Model (continued)},
]
\label{box:annotation-prompt}

\setlength{\parindent}{0pt}
\renewcommand{\baselinestretch}{0.92}\selectfont

\begin{lstlisting}

(*@\textbf{SYSTEM PROMPT}@*)

You are a helpful assistant expert at solving math problems. Please solve the following math problem. First, think through the problem step by step. At the end, output the final answer only in the following format: "Thus, the final answer is: \\boxed{[numeric or mathematical expression only]}"

(*@\textbf{USER PROMPT TEMPLATE}@*)

{{question}}

\end{lstlisting}
\end{tcolorbox}

\subsection{Baselines}
\label{app:baselines}

We evaluate the quality of greedy-pruned reasoning chains in a teacher–pruner–student distillation framework by comparing them with pruning baselines that induce a token-level importance (or an equivalent pruning rule).
We describe each baseline in detail below.

\begin{itemize}[itemsep=1pt, topsep=5pt, leftmargin=*]
\item \texttt{TokenSkip}~\cite{xia-etal-2025-tokenskip}.
\texttt{TokenSkip} prunes reasoning chains using a learned notion of token-level semantic importance.
Similar to our setup, \texttt{TokenSkip} first generates reasoning chains with a teacher model and then prunes them offline.
Token-level importance is predicted by a bidirectional Transformer trained for token-importance estimation~\cite{Pan2024LLMLingua2DD}, using supervision derived from frontier models (e.g., GPT-4~\cite{Achiam2023GPT4TR}). For a token at position $t$, semantic importance is given by the predicted probability of the ``important'' class,
\[
I(t) = P_\phi(y_t = 1 \mid x_{1:T}),
\]
where $P_\phi$ denotes the importance-classification head and $x_{1:T}$ is the full reasoning sequence.
Tokens are ranked by $I(t)$, and those with lower predicted importance are pruned first.
For a fair comparison, we use the same released semantic-importance model (\texttt{llm-lingua-2-xlm-roberta-large}) as in \texttt{TokenSkip} to prune reasoning tokens.

    \item \texttt{H2O}~\cite{Zhang2023H2OHO} is a KV-cache eviction method that identifies ``heavy-hitter'' tokens based on accumulated attention.
Specifically, for a token at position $t$, \texttt{H2O} defines its importance as the cumulative attention it receives from future generation steps,
\[
I(t) = \sum_{i=t+1}^{T} \sum_{l,h} A^{(l,h)}_{i \rightarrow t},
\]
where $A^{(l,h)}_{i \rightarrow t}$ denotes the attention weight from token $i$ to token $t$ at layer $l$ and head $h$.
Tokens with low cumulative attention are evicted from the KV cache. We note that \texttt{H2O} was originally designed for KV-cache eviction rather than explicit token deletion.
Under KV-cache eviction, information associated with a token may still be indirectly accessible via the KV-cache of other tokens, whereas token pruning removes the token entirely from the sequence.
In our experiments, we adapt this notion of importance to token pruning by ranking reasoning tokens according to $I(t)$ and deleting the lowest-ranked tokens.

    \item \texttt{Surprisal}~\cite{li-etal-2023-compressing}.
    Surprisal (self-information) quantifies the information content of a token under a given probability distribution~\cite{Shannon1948AMT}.
    For a token at position $t$, we compute its surprisal under the teacher model using teacher forcing on the full reasoning chain,
\[
I(t) = -\log P(x_t \mid x_{<t}).
\]
    Unlike attention-based importance, surprisal reflects local token predictability rather than task-level or reasoning-specific importance.
    Tokens are ranked by $I(t)$, and those with lower surprisal (higher probability) are pruned first.

  \item \texttt{Uniform}. As a non-informative baseline, we delete reasoning tokens uniformly at random to match the target keep ratio.

\end{itemize}

For fair comparison across methods, we define keep ratios using the student tokenizer and ensure that all pruned reasoning chains contain same number of student-tokenizer tokens.

\subsection{Training and Inference}
\label{app:train_and_inf}

Below, we describe the training and inference implementation details for different stages of our setup:

\begin{itemize}[itemsep=1pt, topsep=5pt, leftmargin=*]

\item \textbf{Reasoning Chain Generation (Teacher).}
We generate reasoning chains via rejection sampling, retaining only responses with correct final answers.
For each question, we sample 10 responses using temperature-based decoding (temperature = 0.7, top-p = 1.0) and randomly select one correct response for training; questions with no correct responses are discarded.
This procedure yields a sufficient number of high-quality reasoning trajectories per dataset for downstream pruning and distillation.
All inference is performed using vLLM.

We construct four effective training datasets for student models under the following teacher–pruner–dataset configurations: (a) Llama3.1-8B as both teacher and pruner on GSM8K; (b) Qwen2.5-7B as both teacher and pruner on GSM8K; (c) Llama3.1-8B as both teacher and pruner on MMLU-Pro; and (d) Qwen2.5-7B as both teacher and pruner on MATH. After reasoning chain generation, the resulting dataset sizes for (a)–(d) are 6,714, 6,650, 5,500, and 3,800 examples, respectively.
We do not use the full training splits listed in Appendix due to the computational cost of pruning and student training, as well as example filtering during rejection sampling.

\item \textbf{Likelihood Computation (Pruner).}
To perform greedy pruning, we represent each question–reasoning–answer instance as a sequence of token IDs and compute post-deletion likelihoods by removing one reasoning token at a time.
All likelihoods are computed under teacher forcing.

We use the vLLM inference engine~\cite{kwon2023efficient}, which supports efficient token-level log-probability evaluation for a fixed input sequence.
For both pruning objectives, we compute token-level log-probabilities under teacher forcing and aggregate them to obtain sequence-level likelihoods.
Specifically, the \textsc{Joint} objective is computed as
\begin{equation}
\scalebox{0.75}{$
\mathcal{L}^{\textsc{Joint}}_{\theta_\Pruner}(Q, R_K, A)
= \sum_{t=1}^{|R_K|+|A|}
\log P_{\theta_\Pruner}\!\left(y_t \mid Q, y_{<t}\right),
$}
\end{equation}
where $\{y_t\}$ denotes the concatenation of the pruned reasoning tokens $R_K$ followed by the answer tokens $A$.
The answer-only objective is computed as
\begin{equation}
\scalebox{0.75}{$
\mathcal{L}^{\textsc{Ans}}_{\theta_\Pruner}(Q, R_K, A)
= \sum_{t=1}^{|A|}
\log P_{\theta_\Pruner}\!\left(a_t \mid Q, R_K, a_{<t}\right).
$}
\end{equation}

During each pruning step, all deletion candidates yield sequences of identical length.
As a result, comparing post-deletion likelihoods using either summed or mean log-probabilities induces the same ranking, and we use the summed formulation throughout.

When pruning a single example, many deletion candidates share long common prefixes, since the input sequences differ only by the removal of a single token.
While vLLM provides automatic prefix caching\footnote{\url{https://docs.vllm.ai/en/stable/features/automatic_prefix_caching/}} for text generation, this functionality does not currently extend to log-probability computation.
Extending prefix caching to likelihood evaluation could substantially reduce the computational cost of greedy pruning.

\item \textbf{Distillation Training (Student).}
All supervised fine-tuning (SFT) experiments are implemented using the \textsc{Axolotl} framework~\cite{axolotl}.
We sweep learning rates in
$\{1\times10^{-6},\,3\times10^{-6},\,5\times10^{-6},\,1\times10^{-5},\,3\times10^{-5},\,5\times10^{-5}\}$
using a cosine decay scheduler.
Across all Llama2-7B training runs, a learning rate of $3\times10^{-5}$ consistently yields the best performance, whereas Mistral-7B prefers a lower learning rate of $5\times10^{-6}$.
All student models are trained with an effective batch size of 28 for 1,200 gradient steps, and the best checkpoint is selected using validation set performance.

\end{itemize}

\subsection{Computational Cost and Scaling of Greedy Pruning}
\label{app:compute_cost}

Greedy pruning incurs a nontrivial computational cost due to repeated likelihood evaluations under token deletions. For a reasoning trace with $n$ tokens and a target keep fraction of $\rho$, the algorithm evaluates $\mathcal{O}((1-\rho)n)$ pruning steps. At each step, all remaining tokens are considered deletion candidates, resulting in a total of approximately $\mathcal{O}((1-\rho)n^2)$ forward passes.
Each candidate evaluation requires scoring the full sequence under the pruner model. Under standard transformer implementations, this corresponds to $\mathcal{O}(n^2)$ self-attention costs per forward pass. Thus, the naive worst-case FLOP complexity is $\mathcal{O}((1-\rho)n^4)$, which can become expensive for longer reasoning traces.

In practice, however, candidate evaluations within each pruning step are independent and can be batched. We implement greedy pruning using \texttt{vLLM} with batched forward passes, so wall-clock time scales with the number of batches rather than with fully sequential execution. Empirically, in our GSM8K setting (average sequence length $\approx 250$ tokens), pruning $8\text{K}$ samples from keep ratio $\rho=1.0 \rightarrow 0.7$ required approximately 20 hours on 8$\times$H100 GPUs. This cost is incurred once as an offline preprocessing step and is amortized over subsequent student training.

\paragraph{Effect of sequence length.}
The quadratic dependence on sequence length implies that greedy pruning becomes increasingly expensive for longer generations (e.g., $>1$K tokens). While our experiments demonstrate feasibility for moderately long reasoning traces (up to $\sim$500 tokens), scaling to substantially longer outputs would require additional approximations.

\paragraph{Optimization opportunities.}
As previously mentioned, several optimizations can reduce the practical runtime:

\begin{itemize}
    \item \textbf{Prefix KV-cache reuse.} For deletion candidates that share common prefixes, key–value states can be reused exactly, reducing redundant computation. While this does not change asymptotic complexity, it yields a meaningful constant-factor speedup.
    
    \item \textbf{Block or local multi-token deletion.} Instead of removing a single token per step, one can remove a small set of high-ranked tokens and recompute importance locally. This reduces the number of pruning iterations at the cost of introducing approximation.
    
    \item \textbf{Surrogate importance models.}~\Cref{subsec:attention-surrogate} shows that attention-based features correlate strongly with pruning ranks. Lightweight surrogate models can approximate token importance and reduce the number of exact likelihood evaluations.
\end{itemize}

\paragraph{Discussion.}
We emphasize that greedy pruning is designed as an \emph{offline analysis and data construction procedure}, rather than a test-time algorithm. Our goal is to expose token-level functional structure and enable efficient distillation, rather than to provide a universally scalable pruning method for arbitrarily long generations. Improving scalability remains an important direction for future work.

\newpage

\section{Additional Distillation Results}
\label{sec:add_dist}

\begin{figure*}[t]
\centering


\small
\legendmark{ub}{}{0pt}~\texttt{$\rho=1.0$}\quad
\legendmark{verbal}{star}{5pt}~\texttt{Greedy}\quad
\legendmark{meta}{isosceles triangle}{6pt}~\texttt{TokenSkip}\quad
\legendmark{entity}{rectangle}{5pt}~\texttt{H2O}\quad
\legendmark{symbolic}{diamond}{5pt}~\texttt{Uniform}\quad
\legendmark{grammar}{regular polygon}{5pt}~\texttt{Surprisal}\quad
\legendmark[dashed]{coref}{}{0pt}~\texttt{ZeroShot}

\vspace{0.5em}

\begin{subfigure}{0.245\textwidth}
  \centering
  \includegraphics[width=\linewidth]{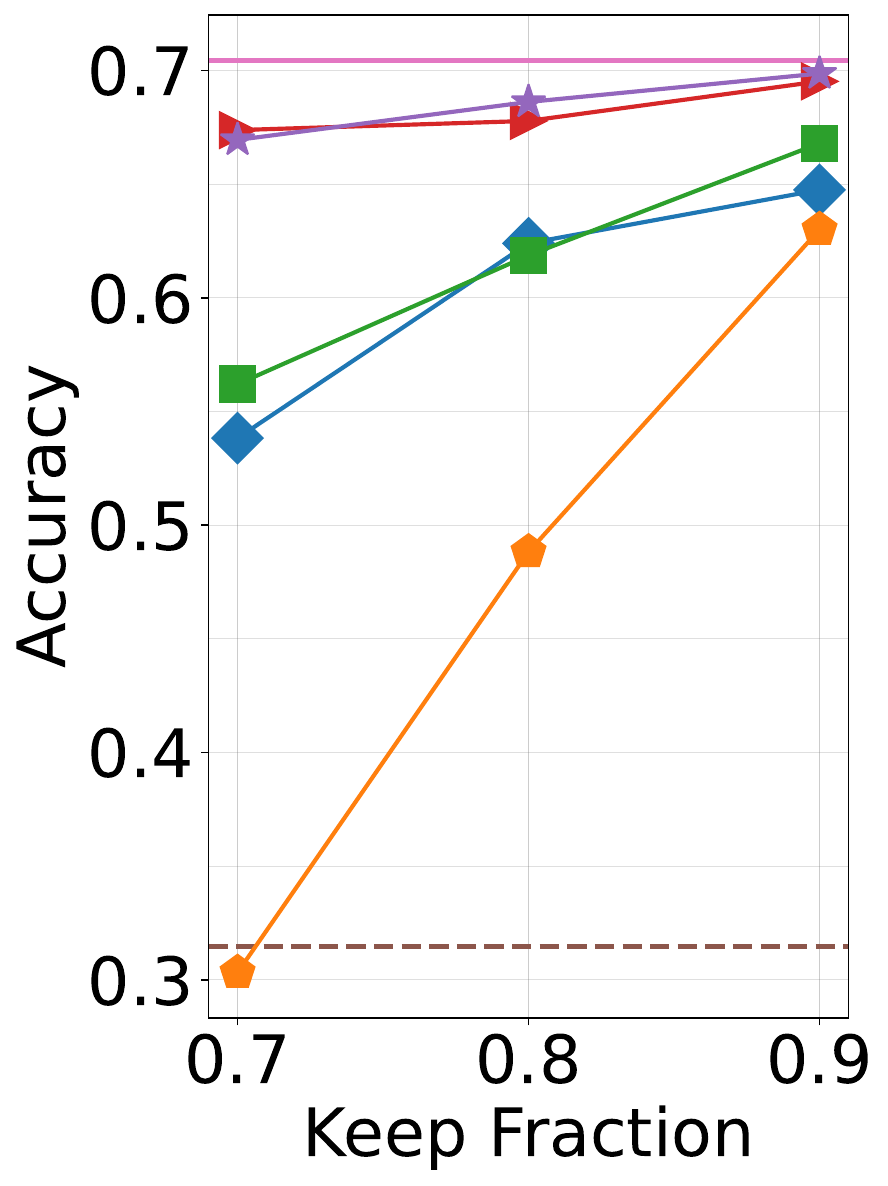}
\caption{{\fontsize{6.8pt}{6.8pt}\selectfont
$\Teacher,\Pruner=\text{Llama3.1-8B}$; GSM8K.}}
\label{fig:mistral-gsm8k-llama3}
\end{subfigure}
\hfill
\begin{subfigure}{0.245\textwidth}
  \centering
  \includegraphics[width=\linewidth]{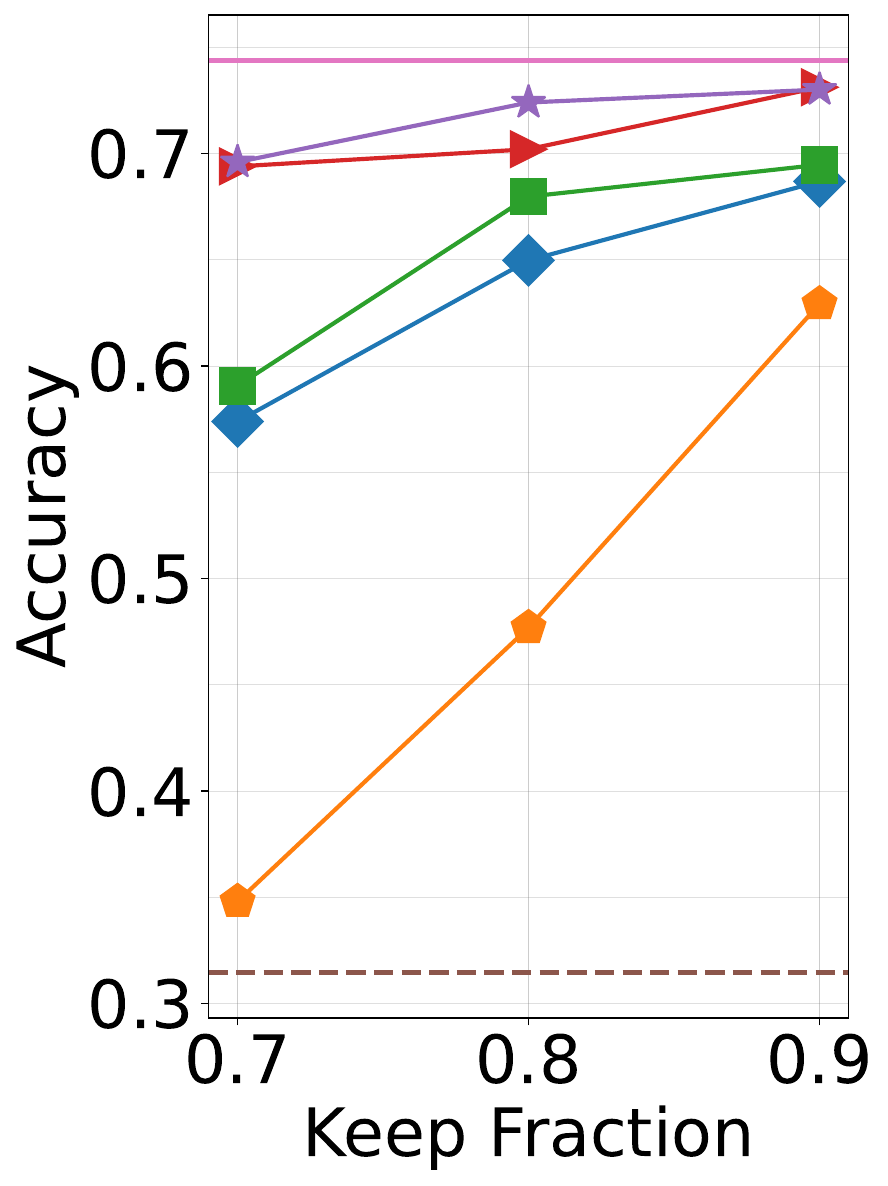}
\caption{{\fontsize{6.8pt}{6.8pt}\selectfont
$\Teacher,\Pruner=\text{Qwen2.5-7B}$; GSM8K.}}
\label{fig:mistral-gsm8k-qwen2.5}
\end{subfigure}
\hfill
\begin{subfigure}{0.245\textwidth}
  \centering
  \includegraphics[width=\linewidth]{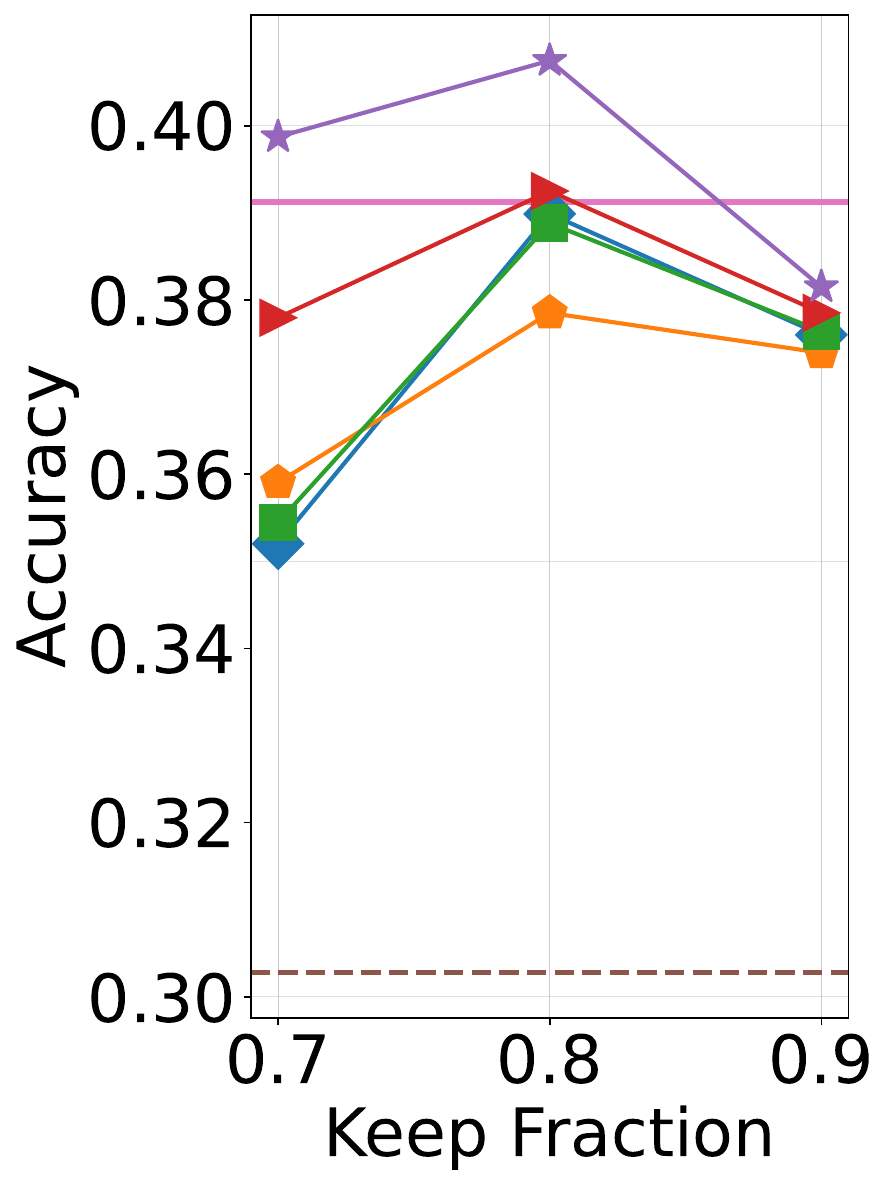}
\caption{{\fontsize{6.7pt}{6.7pt}\selectfont
$\Teacher,\Pruner=\text{Llama3.1-8B}$; MMLU-Pro.}}
\label{fig:mistral-math500-qwen2.5}
\end{subfigure}
\hfill
\begin{subfigure}{0.245\textwidth}
  \centering
  \includegraphics[width=\linewidth]{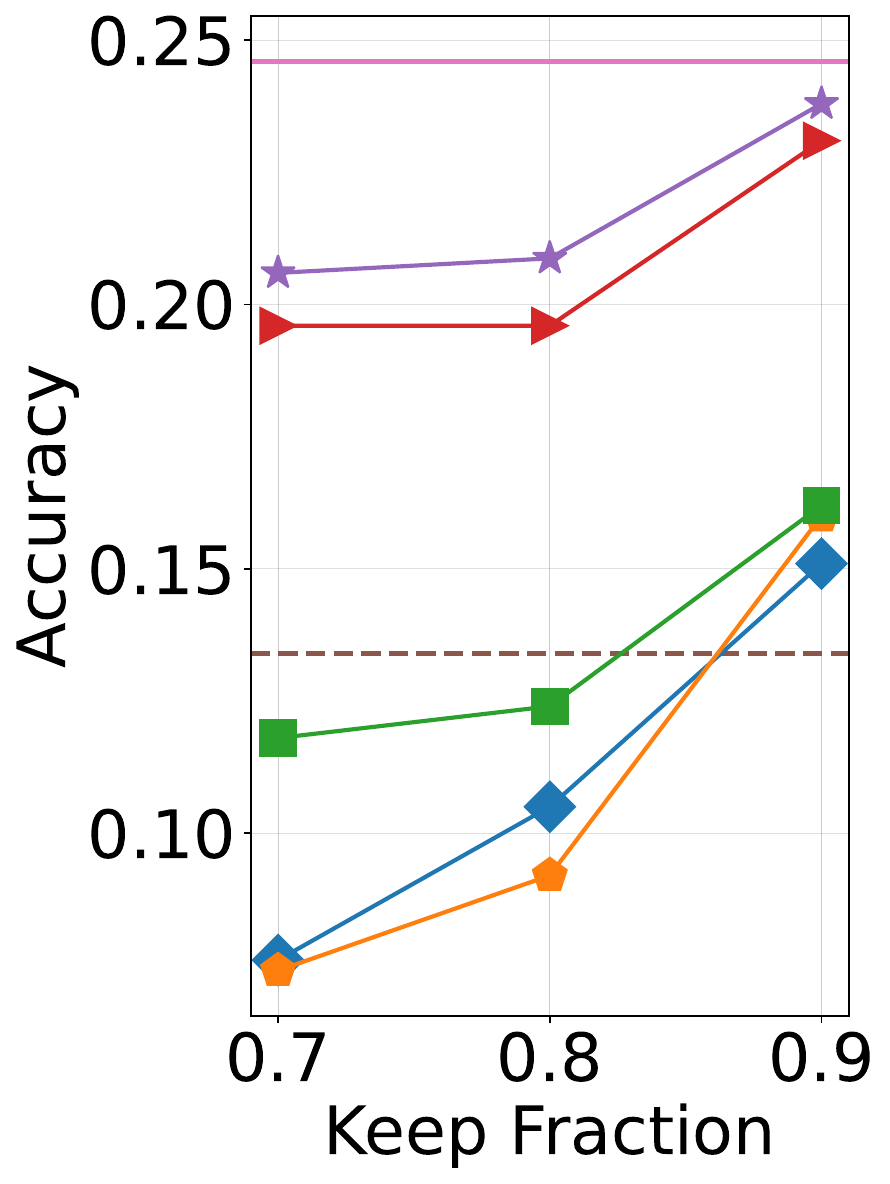}

\caption{{\fontsize{6.8pt}{6.8pt}\selectfont
$\Teacher,\Pruner=\text{Qwen2.5-7B}$; MATH.}}
\label{fig:mistral-mmlu-llama-3}
\end{subfigure}

\caption{
\textbf{Distillation under reasoning token pruning.}
Accuracy of a Mistral-7B student trained on pruned reasoning at varying keep fractions, teacher, pruner, and dataset; dashed lines indicate zero-shot performance.
Greedy pruning achieves the strongest performance at matched lengths, indicating preservation of important tokens.
}
\label{fig:main-results-mistral}
\end{figure*}

We report additional distillation results using Mistral-7B as the student model, shown in Figure~\ref{fig:main-results-mistral}.
Consistent with results using Llama2-7B as the student (Section~\ref{subsec:distill-results}), greedy pruning produces pruned reasoning chains that achieve the strongest distillation performance among baseline methods.
These findings further support the conclusion that greedy pruning preserves functionally important reasoning tokens and indicate that the observed trends generalize across diverse student models.

\newpage

\section{Functional Structure Under Pruning}
\label{sec:appendix-token-category}

\subsection{Categorization Scheme}
\label{sec:appendix-categorization-scheme}

To analyze which tokens of a model’s reasoning are preserved or removed under greedy pruning, we annotate each token in the generated reasoning chain according to its \emph{functional role in reasoning}.
Broadly, we distinguish among tokens that organize or narrate the reasoning process, tokens that support grammatical fluency or referential bookkeeping, and tokens that encode arithmetic content.
Specifically, we assign each token \emph{exactly one} of six categories:

\begin{itemize}[itemsep=1pt, topsep=5pt, leftmargin=*]
  \item \textsc{SymbMath}. \textsc{SymbolicMath} includes explicit numeric or symbolic computation, including digits, currency symbols, arithmetic operators, equations, fractions, and numeric constants.

  \item \textsc{MetaDisc}. \textsc{MetaDiscourse} includes tokens that organize, narrate, or scaffold the reasoning process. Arithmetic verbs (e.g., add, subtract) are labeled \textsc{MetaDiscourse} only when used to narrate a step, not when expressing a mathematical relation. \emph{Examples:} ``to find'', ``we need to'', ``step'', ``first'', ``now'', ``so'', ``let’s'', ``calculate'', ``the final answer is'', step nos., list markers.

  \item \textsc{CoRef}. \textsc{CoReference} includes pronouns and referential expressions that refer to previously introduced entities or quantities. \emph{Examples:} ``she'', ``her'', ``it'', ``they'', ``we'', ``this''.

  \item \textsc{EntName}. \textsc{EntityName} includes proper names and concrete entities central to the problem, including people, objects, units, and counted nouns. \emph{Examples:} ``Natalia'', ``Julie'', ``wallet'', ``book'', ``pages'', ``year''.

  \item \textsc{VerbalMath}. This category includes tokens that describe arithmetic relations or quantities in natural language. Arithmetic verbs (e.g., add, multiply, divide) are labeled \textsc{VerbalMath} only when describing the operation itself, not when narrating a step. \emph{Examples:} ``half'', ``twice'', ``total'', ``remaining'', ``more'', ``per'', ``rate''.

  \item \textsc{Grammar}. This category includes tokens that act as grammatical fillers with minimal standalone content, including articles, prepositions, conjunctions, auxiliary verbs, punctuation, formatting tokens, and whitespace.

\end{itemize}
\noindent All tokens are labeled independently with exactly one category, subject to the constraint that all subword pieces of the same word share the same category. Furthermore, when a token could fit multiple categories, we resolve ties using a fixed priority order to ensure deterministic annotation (\textsc{SymbMath} $>$ \textsc{MetaDisc} $>$ \textsc{CoRef} $>$ \textsc{EntName} $>$ \textsc{VerbalMath} $>$ \textsc{Grammar}).

\subsection{Annotation Process Details}
\label{subsec:annotation_process}
We annotate reasoning tokens with \texttt{gpt-5-mini-2025-08-07} under fixed system and user prompts. 
The system prompt specifies the annotator role and meta-task instructions, while the user prompt defines the annotation task, category definitions, and output constraints; both are included at the end of this section. 
Although \texttt{gpt-5-mini} does not expose sampling settings (e.g., temperature), the labeling procedure is highly stable under a fixed prompt. 
Rerunning the annotation on the same reasoning traces yields a mean token-level agreement of $>$0.90 and a median agreement of 0.92 across 1,000 GSM8K examples, with agreement exceeding 0.97 for 90\% of examples. 
Because our goal is to capture coarse structural trends rather than fine-grained linguistic distinctions, we employ lightweight validation rather than full-scale human annotation. We additionally randomly sample 20 reasoning traces (approximately 6{,}000 tokens) and manually inspect the assigned labels, observing no systematic category collapse or prompt-induced artifacts. Together, these checks indicate that the labeling procedure is sufficiently stable for analyzing relative pruning behavior across functional categories.

\subsection{Additional Results}
\label{sec:appendix-additional-plots}

\begin{figure*}[t]
\centering


\small
\legendmark{symbolic}{circle}{5pt}~\textsc{SymbMath}\quad
\legendmark{grammar}{star}{5pt}~\textsc{Grammar}\quad
\legendmark{entity}{isosceles triangle}{6pt}~\textsc{EntName}\quad
\legendmark{meta}{rectangle}{5pt}~\textsc{MetaDisc}\quad
\legendmark{verbal}{regular polygon}{5pt}~\textsc{VerbalMath}\quad
\legendmark{coref}{diamond}{5pt}~\textsc{CoRef}

\vspace{0.5em}

\begin{subfigure}{0.31\textwidth}
  \centering
  \includegraphics[width=\linewidth]{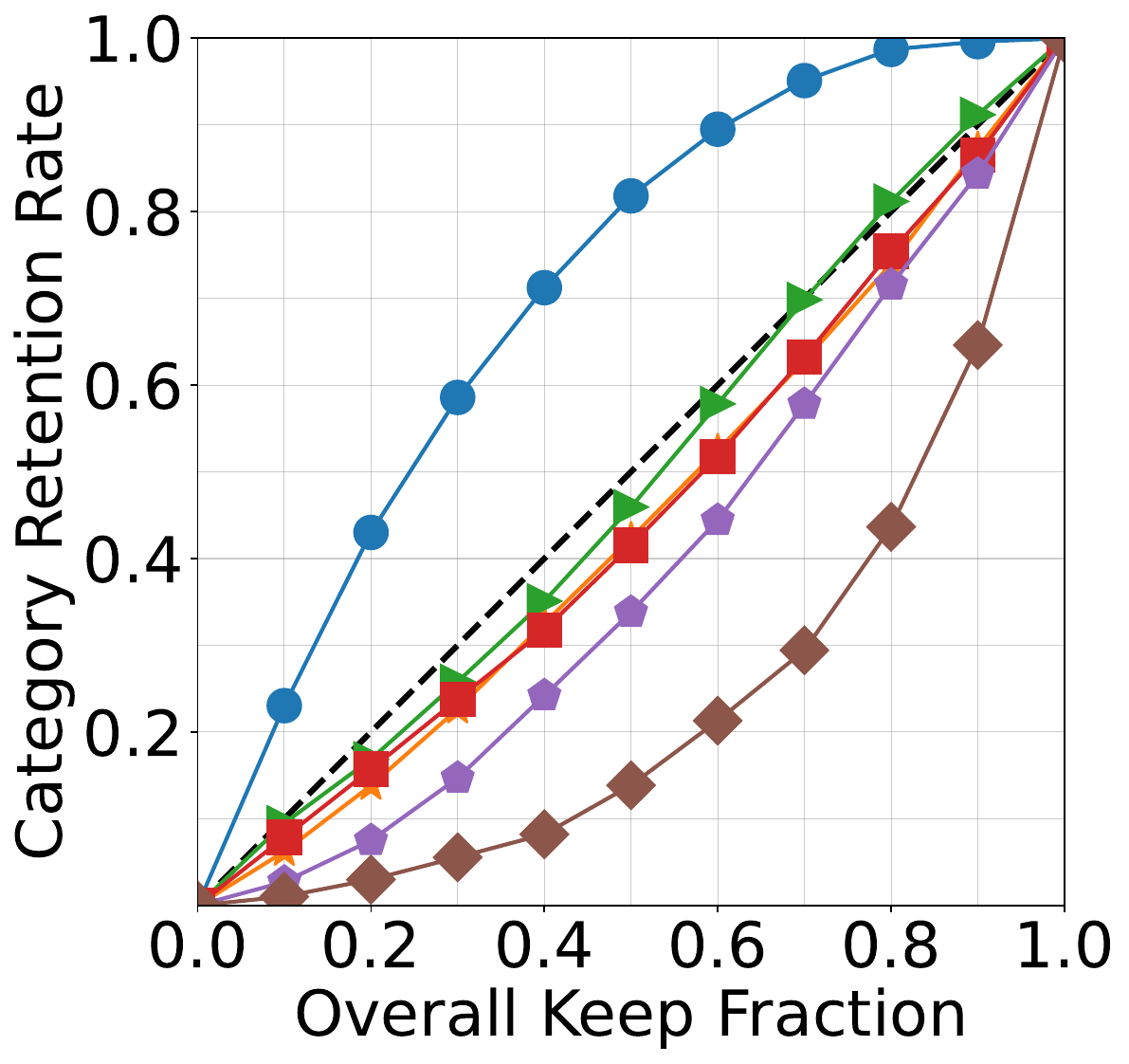}
\caption{{\fontsize{7.0pt}{7.0pt}\selectfont
$\Teacher,\Pruner=\texttt{Llama3.1-8B};\ \mathcal{L}^{\textsc{Joint}}_{\theta_\Pruner}$}}
\label{fig:functional-structure-app-a}
\end{subfigure}
\hfill
\begin{subfigure}{0.31\textwidth}
  \centering
  \includegraphics[width=\linewidth]{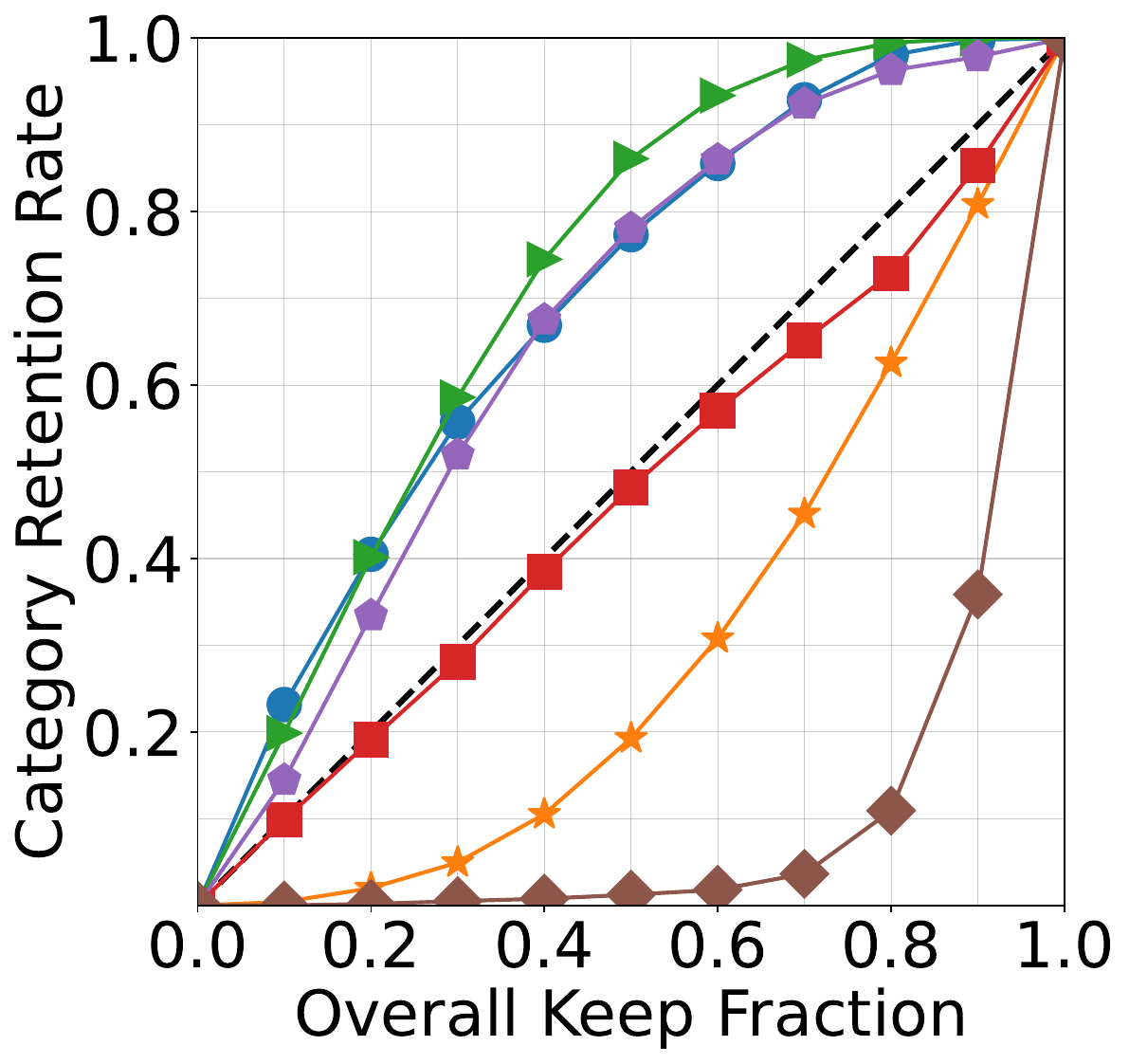}
\caption{{\fontsize{7.0pt}{7.0pt}\selectfont
$\Teacher=\texttt{Qwen2.5-7B};$ \texttt{TokenSkip}}}
\label{fig:functional-structure-app-b}
\end{subfigure}
\hfill
\begin{subfigure}{0.31\textwidth}
  \centering
  \includegraphics[width=\linewidth]{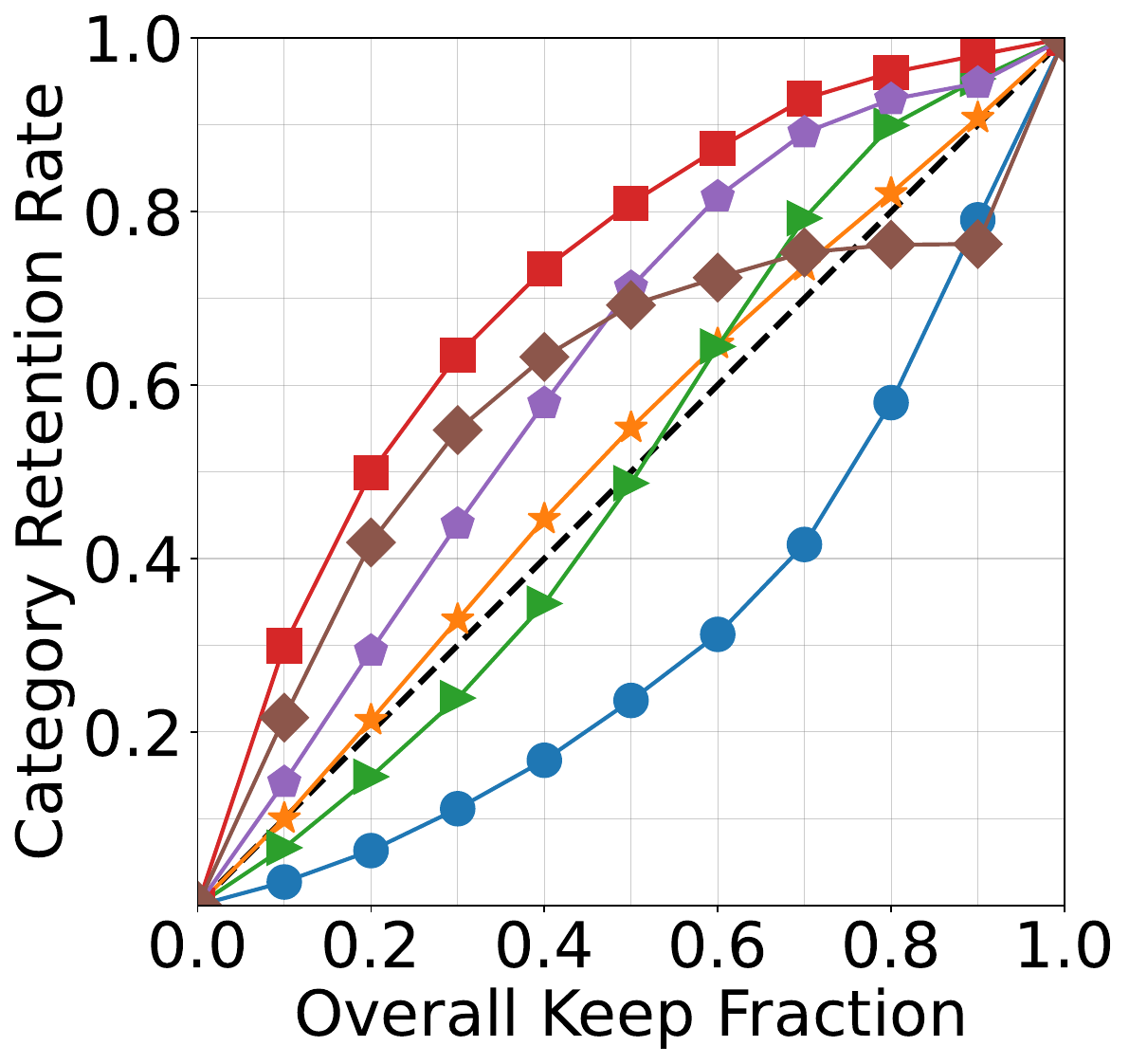}
\caption{{\fontsize{7.0pt}{7.0pt}\selectfont
$\Teacher=\texttt{Qwen2.5-7B};$ \texttt{Surprisal}}}
\label{fig:functional-structure-app-c}
\end{subfigure}

\caption{{
\textbf{Functional structure under different pruning criteria.}
Each curve shows the fraction of tokens retained within a functional category at a given keep fraction; the dashed line denotes uniform pruning.
(a) Greedy pruning with Llama3.1-8B preserves a clear functional ordering, strongly retaining symbolic computation while pruning referential, descriptive, and grammatical scaffolding.
(b) \texttt{TokenSkip} exhibits weaker separation across functional categories, with symbolic computation tracking closer to non-symbolic components and early removal of coreference and grammar.
(c) \texttt{Surprisal}-based pruning under-retains symbolic computation while over-retaining narrative and descriptive categories, yielding a functional structure misaligned with reasoning requirements.
}}
\label{fig:functional-structure-appendix}
\end{figure*}

Figure~\ref{fig:functional-structure-appendix} presents additional functional-structure analyses comparing greedy pruning to two baseline pruning criteria, \texttt{TokenSkip} and \texttt{Surprisal}.
Consistent with the main analysis, \texttt{TokenSkip} exhibits a markedly different functional structure from greedy pruning.
While symbolic content is retained to some extent, \textsc{SymbMath} no longer shows strong preferential retention relative to other categories and instead tracks closer to \textsc{EntName} and \textsc{VerbalMath}.
At the same time, \textsc{Grammar} tokens are pruned aggressively across most keep fractions, and \textsc{CoRef} tokens are removed almost entirely at early stages.
Overall, the retention curves are more tightly clustered, indicating weaker separation across functional roles.
This pattern suggests that semantic-importance signals derived from frontier models emphasize surface-level semantic salience rather than the internal computational structure supporting symbolic reasoning.

\texttt{Surprisal}-based pruning induces an even less aligned functional structure.
High-probability tokens, including \textsc{SymbMath}, are pruned early, leading to under-retention of symbolic computation relative to descriptive categories.
In contrast, \textsc{MetaDisc} and \textsc{VerbalMath} are over-retained, reflecting the tendency of surprisal to preserve less predictable narrative or descriptive tokens rather than computation-critical ones.
\textsc{Grammar} and \textsc{EntName} show inconsistent retention behavior across keep fractions, with patterns largely driven by local token predictability rather than reasoning function.
As a result, \texttt{surprisal} produces a functional structure that is misaligned with the requirements of reasoning.

In contrast, greedy pruning induces a clearer and more interpretable functional ordering, sharply separating symbolic computation from supporting linguistic scaffolding.
These structural differences provide additional context for the weaker distillation performance observed for \texttt{TokenSkip} and \texttt{Surprisal} relative to greedy pruning.

\begin{figure*}[t]
\centering

\newcommand{\legendbox}{\raisebox{0.0ex}{\rule{1.8em}{1.6ex}}}

\small
\textcolor{symbolic}{\legendbox}~\textsc{SymbMath}\quad
\textcolor{grammar}{\legendbox}~\textsc{Grammar}\quad
\textcolor{entity}{\legendbox}~\textsc{EntName}\quad
\textcolor{meta}{\legendbox}~\textsc{MetaDisc}\quad
\textcolor{verbal}{\legendbox}~\textsc{VerbalMath}\quad
\textcolor{coref}{\legendbox}~\textsc{Coref}

\vspace{0.5em}

\makebox[0.85\linewidth]{%
\begin{subfigure}{0.4\linewidth}
    \centering
    \includegraphics[width=\linewidth]{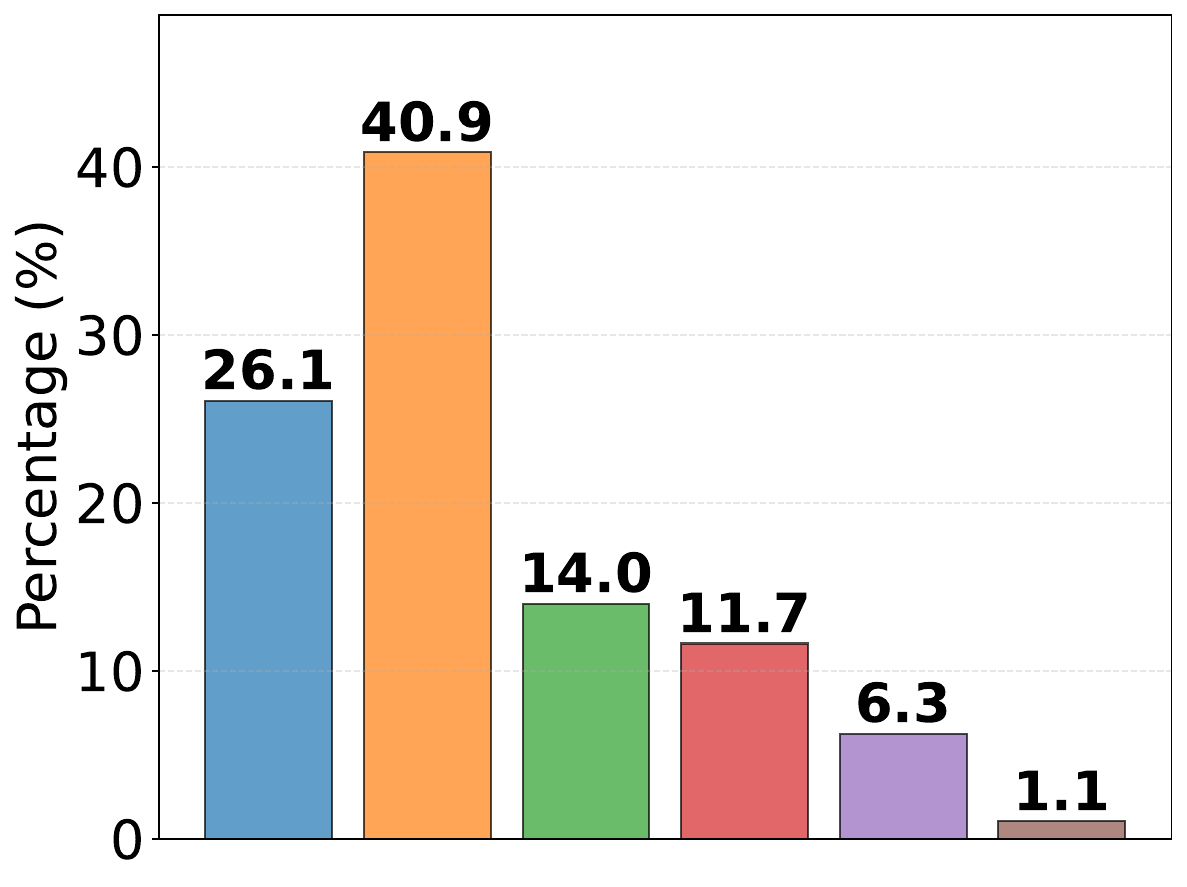}
    \caption{Reasoning generated from \texttt{Qwen2.5-7B}.}
\end{subfigure}
\hfill
\begin{subfigure}{0.4\linewidth}
    \centering
    \includegraphics[width=\linewidth]{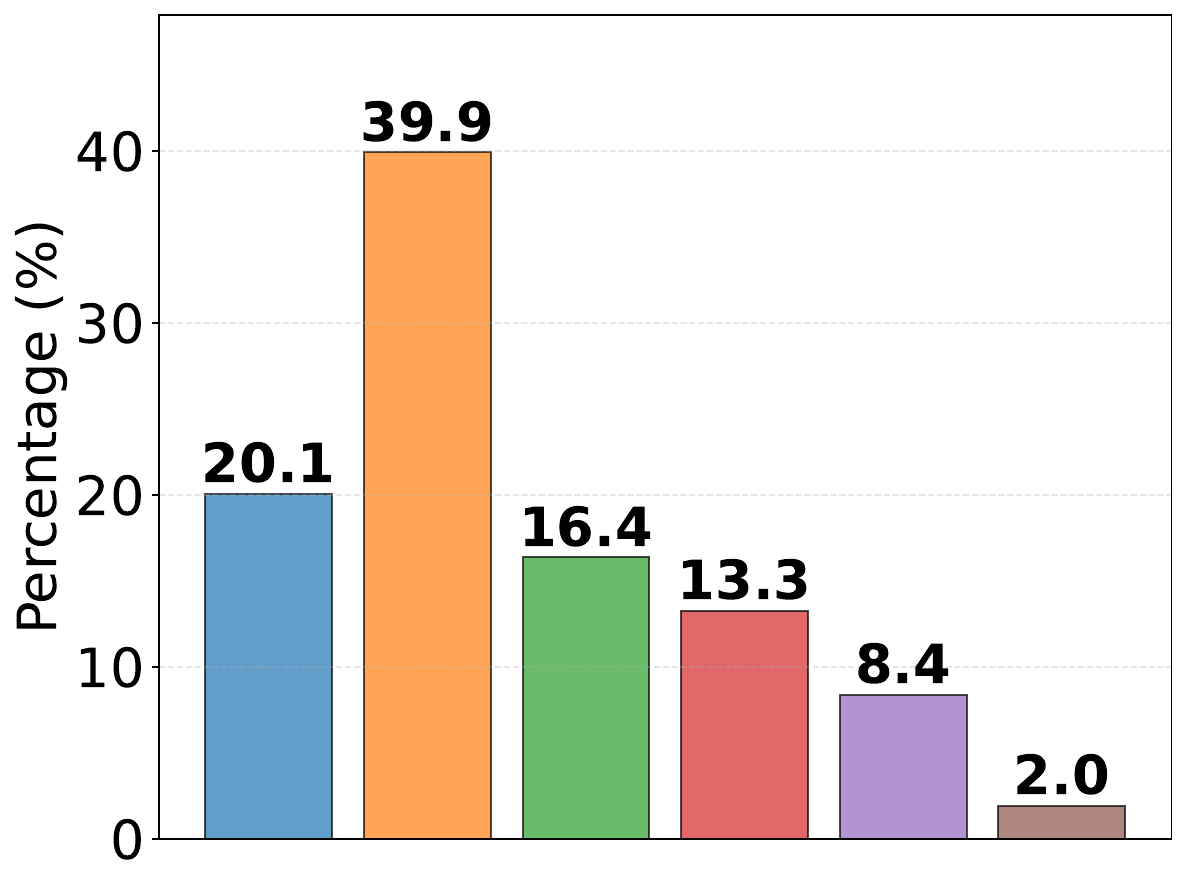}
    \caption{Reasoning generated from \texttt{Llama3.1-8B}.}
\end{subfigure}
}

\caption{
\textbf{Token Category Distribution.}
Token-level functional category distribution over 1,000 randomly sampled GSM8K reasoning traces generated by (a) Qwen2.5-7B and (b) Llama3.1-8B.
Percentages are computed over all reasoning tokens prior to pruning and provide additional context for our functional structure analysis. 
}
\label{fig:cat-dist}
\end{figure*}

Figure~\ref{fig:cat-dist} additionally reports the token-level functional category distribution for reasoning traces generated by Qwen2.5-7B and Llama3.1-8B prior to pruning.
These distributions provide context for the functional-structure analyses by showing the baseline prevalence of each category in unpruned reasoning traces.

\newpage

\section{Comprehensibility and SFT Suitability of Pruned Reasoning}
\label{app:corruption}

We evaluate the structural and semantic quality of reasoning traces produced by greedy pruning. While our method optimizes a likelihood-based objective, it is important to verify that the resulting supervision remains suitable for training and interpretable to humans.

\subsection{Evaluation Protocol}

We evaluate pruned reasoning traces on GSM8K using Llama3.1-8B as the teacher and pruner. For each keep fraction $\rho \in \{0.7, 0.8, 0.9\}$, we sample 1,000 reasoning traces after pruning.

Each trace is evaluated along five dimensions:
\begin{itemize}[noitemsep, topsep=0pt, parsep=0pt, partopsep=0pt]
    \item \textbf{Corruption}: surface-level degradation of text.
    \item \textbf{Structural integrity}: preservation of a stepwise reasoning structure.
    \item \textbf{Semantic coherence}: logical consistency and readability.
    \item \textbf{Mathematical state preservation}: whether key variables, intermediate quantities, and final answers are present and consistent.
    \item \textbf{Suitability for SFT}: whether the trace is usable as supervision for fine-tuning.
\end{itemize}

Annotations are generated using an LLM-based evaluator (temperature $=0$) under a fixed JSON schema to ensure deterministic outputs. To validate annotation quality, we manually inspect 50 randomly sampled traces at $\rho = 0.7$.

\subsection{Annotation Rubric}

Each dimension is labeled using a discrete scale:

\paragraph{Corruption}
\begin{itemize}[noitemsep, topsep=0pt, parsep=0pt, partopsep=0pt]
    \item \textbf{OK}: fully readable text.
    \item \textbf{Minor}: surface artifacts such as missing punctuation, dropped function words, or formatting noise; meaning remains recoverable.
    \item \textbf{Severe}: fragmented or malformed text that is difficult to parse.
\end{itemize}

\paragraph{Structural Integrity}
\begin{itemize}[noitemsep, topsep=0pt, parsep=0pt, partopsep=0pt]
    \item \textbf{Intact}: step structure preserved.
    \item \textbf{Mildly broken}: minor step merging or header loss; order remains recoverable.
    \item \textbf{Broken}: reasoning flow is difficult to reconstruct.
\end{itemize}

\paragraph{Semantic Coherence}
\begin{itemize}[noitemsep, topsep=0pt, parsep=0pt, partopsep=0pt]
    \item \textbf{OK}: logically understandable.
    \item \textbf{Minor}: connective or phrasing gaps, but reasoning remains clear.
    \item \textbf{Severe}: logical flow is disrupted.
\end{itemize}

\paragraph{Mathematical State Preservation}
\begin{itemize}[noitemsep, topsep=0pt, parsep=0pt, partopsep=0pt]
    \item \textbf{Yes}: key quantities and equations are present and numerically consistent.
    \item \textbf{Borderline}: minor omissions such as units; state remains inferable.
    \item \textbf{No}: critical quantities are missing or inconsistent.
\end{itemize}

\paragraph{Suitability for SFT}
\begin{itemize}[noitemsep, topsep=0pt, parsep=0pt, partopsep=0pt]
    \item \textbf{Yes}: directly usable supervision.
    \item \textbf{Borderline}: usable with minor cleanup.
    \item \textbf{No}: too degraded for training.
\end{itemize}

\subsection{Results}

Table~\ref{tab:corruption_full} reports results across keep fractions.

\begin{table}[h]
\centering
\begin{adjustbox}{width=0.9\linewidth}
\begin{tabular}{llccc}
\toprule
\textbf{Metric} & \textbf{Category} & $\rho = 0.7$ & $\rho = 0.8$ & $\rho = 0.9$ \\
\midrule

\multirow{3}{*}{Corruption}
& OK        & 3.9\%  & 5.1\%  & 22.7\% \\
& Minor     & 96.1\% & 94.9\% & 77.3\% \\
& Severe    & 0.0\%  & 0.0\%  & 0.0\% \\

\midrule

\multirow{3}{*}{Structural Integrity}
& Intact        & 57.3\% & 78.8\% & 92.2\% \\
& Mildly broken & 42.4\% & 21.2\% & 7.8\% \\
& Broken        & 0.4\%  & 0.0\%  & 0.0\% \\

\midrule

\multirow{3}{*}{Semantic Coherence}
& OK     & 62.0\% & 76.1\% & 92.2\% \\
& Minor  & 38.0\% & 23.9\% & 7.8\% \\
& Severe & 0.0\%  & 0.0\%  & 0.0\% \\

\midrule

\multirow{3}{*}{Math State Preserved}
& Yes        & 94.5\% & 97.3\% & 99.6\% \\
& Borderline & 5.5\%  & 2.7\%  & 0.4\% \\
& No         & 0.0\%  & 0.0\%  & 0.0\% \\

\midrule

\multirow{3}{*}{SFT Suitability}
& Yes        & 67.1\% & 85.9\% & 94.1\% \\
& Borderline & 32.9\% & 14.1\% & 5.9\% \\
& No         & 0.0\%  & 0.0\%  & 0.0\% \\

\bottomrule
\end{tabular}
\end{adjustbox}
\caption{Evaluation of pruned reasoning traces across multiple quality dimensions and keep fractions $\rho$.}
\label{tab:corruption_full}
\end{table}

\subsection{Analysis}

At $\rho = 0.7$ (approximately 30\% token reduction), degradation is primarily superficial. Most traces exhibit minor corruption, such as missing punctuation or dropped function words, but remain semantically coherent and structurally interpretable. Severe corruption is absent across all settings. Structural failures are rare (0.4\% at $\rho = 0.7$), and no examples exhibit a loss of mathematical state. In all cases, final answers remain present and numerically consistent. Suitability for SFT remains high even at aggressive pruning levels. At $\rho = 0.7$, 67.1\% of traces are directly usable, and the remaining ones are borderline, requiring only minor cleanup. No traces are deemed unusable. Manual inspection confirms that degradation is dominated by formatting artifacts rather than logical errors. Key reasoning steps, intermediate quantities, and final answers are preserved.

\paragraph{Takeaways:}
These results suggest that greedy pruning preserves the semantic and mathematical integrity of reasoning traces even under substantial compression. While surface fluency degrades at lower keep fractions, this effect is largely orthogonal to the underlying reasoning and can be addressed with lightweight post-processing if desired. Overall, pruned traces remain suitable supervision for distillation and support the use of greedy pruning as a token-level compression method.

\begin{tcolorbox}[
  breakable,
  colback=white,
  colframe=black,
  title={Prompt: Corruption Annotation},
  label={box:corruption-annotation-prompt},
  title after break={Prompt: Corruption Annotation (continued)},
]

\setlength{\parindent}{0pt}
\renewcommand{\baselinestretch}{0.92}\selectfont

\begin{lstlisting}

(*@\textbf{SYSTEM PROMPT}@*)

You are a strict annotator, evaluating the corruption of a pruned response compared to the original response.

(*@\textbf{USER PROMPT TEMPLATE}@*)

You are evaluating whether a pruned response is corrupted, using the following rubric.

Your task has TWO STEPS, to be performed and output in order:

Step 1 (Reasoning): Compare the PRUNED response to the ORIGINAL response step by step, in order.
- For each step, directly compare the pruned and full response step (first with first, second with second, etc).
- For each pair of steps, write a single line of analysis that considers: corruption (surface-level preservation, missing words, fragments, readability), structural integrity (step structure), semantic coherence (clarity of reasoning), math state (clarity/preservation of quantities), and suitability for SFT.
- For each step, clearly state what is intact, what is mildly broken, and what is severely degraded according to the rubric.
- Do not evaluate answer correctness; focus only on surface damage, structure, and coherence.
- Write one line of analysis per step, in order, each on its own NEW LINE, precisely covering all rubric metrics for the step-by-step differences.

Step 2 (Label): On a NEW LINE, output ONLY the final JSON (no extra text before or after it), using EXACTLY this schema and label space:

{
  "corruption": "ok|minor|severe",
  "structural_integrity": "intact|mildly_broken|broken",
  "semantic_coherence": "ok|minor|severe",
  "math_state_preserved": "yes|borderline|no",
  "suitable_for_sft": "yes|borderline|no",
  "notes": ""
}

Rubric (for labeling):
- corruption:
    * ok (readable/coherent),
    * minor (surface artifacts such as missing punctuation or truncated fragments but meaning recoverable),
    * severe (hard to parse or nonsensical)
- structural_integrity: intact / mildly_broken / broken (whether step structure survives)
- semantic_coherence: ok / minor / severe (whether the reasoning remains understandable to a human reader)
- math_state_preserved: yes (main intermediate quantities/equations clearly present and consistent) / borderline (some degradation) / no (missing or inconsistent)
- suitable_for_sft: yes (readable and structurally usable for SFT), borderline (usable but mildly degraded), no (too broken for practical SFT)
- notes: If everything is perfect (corruption="ok" AND structural_integrity="intact" AND semantic_coherence="ok"), notes MUST be "". If suitable_for_sft="no", notes may still be "". Otherwise, use <= 15 words describing the surface-level defects.

Remember:
- Do NOT evaluate correctness of the answer.
- Assess only readability, structure, and coherence of the PRUNED text versus the FULL text.

Question:
{{question}}

Original Response (full):
{{reason}}

Pruned Response:
{{pruned_reason}}

First think step-by-step about each corresponding step in the pruned response and the original response. Then return the JSON. Return EXACTLY this JSON schema:
{
  "corruption": "ok|minor|severe",
  "structural_integrity": "intact|mildly_broken|broken",
  "semantic_coherence": "ok|minor|severe",
  "math_state_preserved": "yes|no|unclear",
  "suitable_for_sft": "yes|borderline|no",
  "notes": ""
}

\end{lstlisting}
\end{tcolorbox}

\newpage

\section{Student Behavior Under Pruned-Trace Distillation}
\label{app:student_behavior}

In this section, we analyze how student models behave after training on greedily pruned reasoning traces. In particular, we study whether (i) students learn to generate shorter reasoning chains consistent with the pruned supervision, and (ii) whether this compression leads to degradation in structural integrity or semantic coherence.

\subsection{Reasoning Length Adaptation}

We first measure the length of reasoning traces generated by student models trained with different keep fractions $\rho \in \{1.0, 0.9, 0.8, 0.7\}$. Table~\ref{tab:student_length_inline} reports the average number of generated reasoning tokens on the GSM8K validation set.

\begin{table}[h]
\centering
\small
\begin{tabular}{c|l}
\toprule
Keep Fraction $\rho$ & Student Output Length \\
\midrule
1.0 & 189 tokens \\
0.9 & 161 tokens ($\sim$0.85$\times$ of $\rho=1.0$) \\
0.8 & 148 tokens ($\sim$0.78$\times$ of $\rho=1.0$) \\
0.7 & 134 tokens ($\sim$0.70$\times$ of $\rho=1.0$) \\
\bottomrule
\end{tabular}
\caption{
Student reasoning lengths under different pruning levels. 
Values in parentheses indicate relative length normalized to the $\rho=1.0$ setting.
}
\label{tab:student_length_inline}
\end{table}

We observe that the student model closely matches the length distribution of the pruned training data. In particular, a keep fraction of $\rho=0.7$ yields approximately $30\%$ shorter reasoning traces at inference time. This indicates that the student does not revert to verbose reasoning, but instead internalizes the compressed reasoning style present in the supervision.

\subsection{Structural and Semantic Properties of Student Outputs}

We next evaluate whether compressed reasoning leads to degradation in output quality. Following the same protocol used for evaluating pruned training traces, we assess student-generated outputs along four dimensions: (i) structural integrity, (ii) semantic coherence, (iii) preservation of mathematical state, and (iv) overall suitability for SFT-style reasoning.

Across all pruning levels, we find that student outputs largely preserve the structure and semantics of the reasoning process. In particular, even at $\rho=0.7$, we do not observe cases where the reasoning becomes logically invalid or mathematically inconsistent. Instead, degradation is primarily stylistic, consisting of:
\begin{itemize}[noitemsep, topsep=0pt, parsep=0pt, partopsep=0pt]
    \item missing function words (e.g., articles, prepositions),
    \item reduced punctuation,
    \item occasional merging of reasoning steps.
\end{itemize}

These artifacts are consistent with those observed in the pruned training data and suggest that the student faithfully learns the distribution of compressed reasoning traces.

\subsection{Interpretability of Compressed Reasoning}

Despite the reduction in length, student-generated reasoning remains interpretable. In most cases, key intermediate quantities and computation steps are preserved, allowing the reasoning process to be followed with minimal effort. The primary effect of pruning is the removal of redundant narrative scaffolding rather than core computational content.

We emphasize that the goal of greedy pruning is not to produce stylistically optimal reasoning, but to preserve functional correctness under compression. If surface fluency is desired, this can be addressed orthogonally (e.g., via a lightweight rewriting or post-processing step), without altering the underlying token importance structure.

\subsection{Summary}

Overall, compressed-reasoning distillation yields student models that:
\begin{itemize}[noitemsep, topsep=0pt, parsep=0pt, partopsep=0pt]
    \item generate expected shorter reasoning traces,
    \item preserve semantics and mathematical state,
    \item exhibit only minor stylistic degradation.
\end{itemize}

These results suggest that greedy pruning provides a viable mechanism for training token-efficient reasoning models without materially compromising interpretability or correctness.

\onecolumn
\begin{tcolorbox}[
  breakable,
  colback=white,
  colframe=black,
  title={Prompt: Functional Role Annotation},
  label={temp},
  title after break={Prompt: Functional Role Annotation (continued)},
]
\label{box:annotation-prompt}

\setlength{\parindent}{0pt}
\renewcommand{\baselinestretch}{0.92}\selectfont

\begin{lstlisting}

(*@\textbf{SYSTEM PROMPT}@*)

You are a careful linguistic annotator. Your task is to label tokens using a fixed category set and strict priority rules. You must follow the definitions exactly and output valid JSON only. Do not include any other text in your response.

(*@\textbf{USER PROMPT TEMPLATE}@*)

You are annotating tokens from a large language model's chain-of-thought according to their FUNCTIONAL ROLE in mathematical reasoning.

Your goal is to assign EXACTLY ONE category per token such that the resulting labels reveal which types of reasoning content are present and how they differ in importance.

Tokens may be subword pieces (e.g., "Weng" -> "W", "eng"; "babysitting" -> "babys", "itting"). Label each token independently, but ALL subword pieces belonging to the same word MUST receive the SAME category.

You MUST choose exactly one of the following categories:

--------------------------------------------------
CATEGORIES
--------------------------------------------------

1. SYMBOLIC_MATH
   - Explicit numeric or symbolic computation
   - Includes: digits, currency symbols, arithmetic operators, equations, fractions, numeric constants
   
2. META_DISCOURSE
   - Tokens that organize, narrate, or scaffold the reasoning process
   - Includes instructional or planning language rather than mathematical content
   - Examples: "to find", "we need to", "step", "first", "now", "so", "let's", "calculate", "find", "the final answer is", step numbers, list markers
   - Arithmetic verbs (e.g., "add", "subtract") are META_DISCOURSE ONLY when they introduce or narrate a step, not when expressing a mathematical relation

3. COREFERENCE
   - Pronouns or references pointing to previously mentioned entities or quantities
   - Includes: she, her, he, it, they, we, him, this
   - "this" is COREFERENCE ONLY when referential, not when used as a grammatical filler

4. ENTITY_NAME
   - Proper names or concrete entities central to the problem
   - Includes: people's names, objects, units, or concrete nouns being counted
   - Examples: "Natalia", "Julie", "wallet", "book", "pages", "year"

5. VERBAL_MATH
   - Natural-language descriptions of arithmetic relationships or quantities
   - Includes: "half", "twice", "total", "remaining", "more", "per", "rate"
   - Arithmetic verbs (e.g., "add", "multiply", "divide") belong here ONLY when describing the operation itself, not when narrating steps

6. GRAMMATICAL
   - Grammatical glue with little standalone semantic content
   - Includes: articles, prepositions, conjunctions, auxiliary verbs, punctuation, formatting tokens, whitespace

--------------------------------------------------
IMPORTANT NOTES
--------------------------------------------------

1. Adjectives are NOT a separate category. Assign adjectives based on function:
  - Arithmetic-modifying adjectives -> VERBAL_MATH
  - Discourse or narrative adjectives -> META_DISCOURSE
  - Entity-identifying adjectives -> ENTITY_NAME
  - Otherwise -> FUNCTION

2. ENTITY_NAME vs VERBAL_MATH: When a noun denotes a concrete object being counted (e.g., "pages", "wallet"), label it ENTITY_NAME. Mathematical relations involving those nouns are captured by VERBAL_MATH or SYMBOLIC_MATH.

--------------------------------------------------
PRIORITY RULES (STRICT)
--------------------------------------------------

If a token could belong to multiple categories, apply the FIRST matching rule:

1. If part of an explicit numeric or symbolic expression -> SYMBOLIC_MATH
2. Else if it narrates or structures reasoning -> META_DISCOURSE
3. Else if it is referential -> COREFERENCE
4. Else if it names a concrete entity -> ENTITY_NAME
5. Else if it describes arithmetic verbally -> VERBAL_MATH
6. Else -> GRAMMATICAL


--------------------------------------------------
CONSISTENCY CONSTRAINT
--------------------------------------------------

If the same surface word appears multiple times with the same functional role, assign it the SAME category across occurrences unless its role clearly changes.

--------------------------------------------------
INPUT FORMAT
--------------------------------------------------
Question:
<question in natural language>

Output Reason:
<reason in natural language>

Output Answer:
<answer in natural language>

Reason Tokens:
<reason tokens in JSON format>
[
  {
    "token_position": <int>,
    "token_text": "<string>",
  }
]

--------------------------------------------------
OUTPUT FORMAT
--------------------------------------------------

Return a JSON list. Each entry MUST have:

{
  "token_position": <int>,
  "token_text": "<string>",
  "category": "<one of the 6 categories>",
  "justification": "<6-word explanation>"
}

Do NOT assign multiple categories.
Do NOT invent new categories.
Be consistent across similar tokens.

--------------------------------------------------
ANNOTATE THE FOLLOWING
--------------------------------------------------

Question:
{{question}}

Output Reason:
{{reason}}

Output Answer:
{{answer}}

Output Reason Tokens (JSON Format):
{{reason_tokens}}

\end{lstlisting}

\end{tcolorbox}
\twocolumn



\end{document}